\documentclass[10pt,twocolumn,letterpaper]{article}

\usepackage[pagenumbers]{cvpr} 

\definecolor{cvprblue}{rgb}{0.21,0.49,0.74}
\usepackage[pagebackref,breaklinks,colorlinks,allcolors=cvprblue]{hyperref}
\usepackage{multirow}
\def\paperID{30135} 
\def\confName{CVPR}
\def\confYear{2026}

\title{Unveiling Modality Bias: Automated Sample-Specific Analysis for Multimodal Misinformation Benchmarks}

\author{Hehai Lin$^{1}$, Hui Liu$^{2}$, Shilei Cao$^{3}$, Jing Li$^{4}$, Haoliang Li$^{2}$, Wenya Wang$^{5}$\thanks{Corresponding to \href{mailto:wangwy@ntu.edu.sg}{wangwy@ntu.edu.sg}}\\
$^{1}$The Hong Kong University of Science and Technology (Guangzhou), \\
$^{2}$City University of Hong Kong, 
$^{3}$Sun Yat-sen University, \\
$^{4}$Harbin Institute of Technology, 
$^{5}$Nanyang Technological University
}

\begin{document}
\maketitle
\begin{abstract}
Numerous multimodal misinformation benchmarks exhibit bias toward specific modalities, allowing detectors to make predictions based solely on one modality.
While previous research has quantified bias at the dataset level or manually identified spurious correlations between modalities and labels, these approaches lack meaningful insights at the sample level and struggle to scale to the vast amount of online information.
In this paper, we investigate the design for automated recognition of modality bias at the sample level.
Specifically, we propose three bias quantification methods based on theories/views of different levels of granularity: 1) a coarse-grained evaluation of modality benefit; 2) a medium-grained quantification of information flow; and 3) a fine-grained causality analysis. 
To verify the effectiveness, we conduct a human evaluation on two popular benchmarks.
Experimental results reveal three interesting findings that provide potential direction toward future research: 1)~Ensembling multiple views is crucial for reliable automated analysis; 2)~Automated analysis is prone to detector-induced fluctuations; and 3)~Different views produce a higher agreement on modality-balanced samples but diverge on biased ones. 
\end{abstract}    
\section{Introduction}
\label{sec:intro}

The proliferation of online social media has accelerated the dissemination of misinformation~\cite{li2024mcfend, bu2024fakingrecipe, wang2024explainable, yue2024evidence, wan2024risk}, particularly in multimodal contexts where images and texts mutually reinforce each other, enhancing persuasiveness and deception to people~\cite{tahmasebi2024multimodal, guo2024each, chen2023combating, comito2023multimodal}.
Such multimodal misinformation poses significant threats to social cohesion~\cite{duffy2020too}, public safety~\cite{roozenbeek2020susceptibility}, and political health~\cite{gamir2021multimodal}.
To verify the ability of Multimodal Misinformation Detection (MMD) models to exploit multimodal information, previous studies have proposed several Multimodal Misinformation Benchmarks (MMBs) such as Fakeddit~\cite{nakamura2019r}
and MMFakeBench~\cite{liu2024mmfakebench},
which have advanced the development of Multimodal Misinformation Detection (MMD). 

\begin{figure*}[t]
  \centering
  \includegraphics[width=\linewidth]{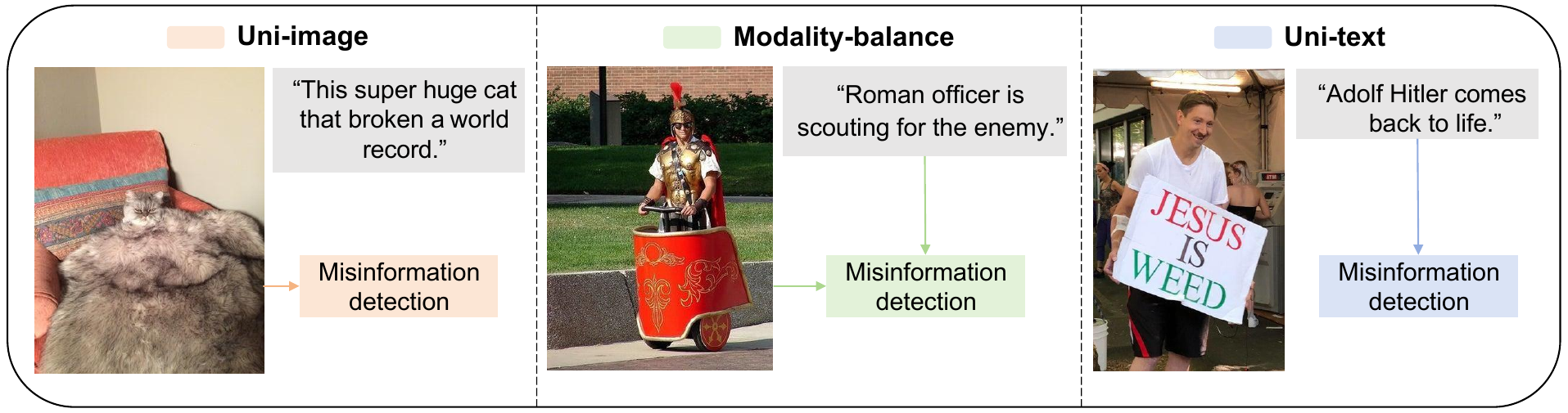}
  \caption{Effect of different samples on misinformation detection. For biased samples, we can directly infer from preferred modality, like the \textbf{Left} (an unreasonable fat cat image) and \textbf{Right} (the impossibility of resurrection) one.}
  \label{fig: background}
\end{figure*}

However, these benchmarks exhibit bias toward specific modality~\cite{papadopoulos2024verite}, where one modality may dominate as the primary source of information, thereby diminishing the role of the other modality~\cite{guo2023modality, liang2024quantifying}. 
For example, it has been observed that the COSMOS benchmark~\cite{aneja2021cosmos} exhibits a text bias, where text-only methods can outperform their multimodal counterpart~\cite{papadopoulos2023synthetic}. 
As shown in Figure~\ref{fig: background}, we can make a judgment by analyzing only the image of a fat cat or the text ``Adolf Hitler comes back to life'' without considering the other modality.
Such modality bias can lead to serious problems: 
First, from the training aspect, models trained on biased benchmarks may lack robustness to the variation of that modality~\cite{yang2024quantifying}, making them vulnerable to uni-modal attacks.
Second, from the evaluation aspect, biased benchmarks may yield an incomprehensive measurement of MMD models, e.g., a model might perform well on a text-biased benchmark because it learns spurious text-label correlations instead of effectively integrating multimodal information~\cite{goyal2017making}.

Unfortunately, no systematic investigation has been conducted on the modality bias of existing MMBs. Current methods for detecting modality bias on general multimodal benchmarks like visual question answering can be broadly divided into two categories: automated dataset-level quantification and manual identification by human experts. For the former one, \citet{liang2024quantifying} utilize information theory to measure \textit{redundancy}, \textit{uniqueness}, and \textit{synergy} across the entire dataset. However, as illustrated in Figure~\ref{fig: background}, bias can vary significantly across individual samples within a dataset, suggesting that this approach lacks the granularity needed to fully capture sample-specific biases. The latter one, as demonstrated by \citet{liu2024tackling}, involves detecting specific issues, such as spurious correlations between text modalities and labels. 
While manual identification can effectively detect biased samples, it is limited by scalability and is impractical for handling a large volume of online data. This naturally raises the question: \textbf{is it possible to automatically measure the modality bias at the sample level without human intervention?}

To this end, we conduct a systematic analysis of modality bias in MMBs and verify whether machines can automatically provide a reasonable measurement.
Modality bias can be classified into three types: Uni-image, Uni-text, and Modality-balance, which indicate image bias, text bias, and no bias.
We leverage three quantification methods of different granularities and adapt them to bias identification, i.e., modality benefit, modality flow, and modality causal effect.
At a coarse level, \textbf{modality benefit} identifies the input modality that contributes the most for final predictions using Shapley values~\cite{shapley1953value, parcalabescu2023mm} from game theory, which fairly assesses individual contributions of different players in cooperative scenarios. 
At a medium level, \textbf{modality flow} utilizes saliency scores~\cite{michel2019sixteen, wang2023label}, which quantify attention interactions between different input modalities and output predictions to inspect the decision-making process.
At the finest level, \textbf{modality causal effect} constructs the causal inference graph of MMD, which contains modality-balanced and biased paths, and traces the path that has the maximal causal effect based on counterfactual reasoning~\cite{chen2023causal, chen2024quantifying}.
We treat these methods as providing different views upon the decision of modality bias and adopt a voting mechanism to integrate these three views to obtain an ensembled multi-view output.
The multi-view analysis is obtained through an ensemble voting mechanism.

To validate the effectiveness of such automated sample-specific bias analysis, we conduct a human evaluation on 300 samples of Fakeddit~\cite{nakamura2019r} and MMFakeBench~\cite{liu2024mmfakebench}.
Experimental results reveal three interesting findings that offer potential direction and design considerations toward future automated sample-specific modality bias analysis: 
1)~Ensembling multiple views is crucial for a reliable automated analysis,
because the intricate nature of automated sample-specific modality bias detection is a complex task for machines. 2)~Automated analysis is prone to detector-induced fluctuations. The performance of both single- and multi-view analysis is sensitive to the choice of misinformation detectors. This phenomenon is unavoidable since each view is dependent on the parameters of the chosen detector.  Mitigating such sensitivity could enhance its practicality for real-world deployment. 3)~Different views produce a higher agreement on modality-balanced samples but diverge on biased ones. We take it reasonable since different views possess distinct patterns for capturing bias. 
Overall, we believe that automated sample-specific analysis has significant practical applications, e.g., cleaning a biased MMB by retaining modality-balanced samples with high consistency.
Our contributions are as follows:
\textbf{Firstly}, we are the first to design an automated sample-specific modality bias analysis for multimodal misinformation benchmarks.
\textbf{Secondly}, we investigate the effectiveness of the proposed automated analysis via a human evaluation on two multimodal misinformation benchmarks.
\textbf{Thirdly}, we uncover some interesting findings from empirical experiments, offering potential directions toward future research. 
\section{Related Work}
\subsection{Modality Bias}
Modality bias is prevalent in various multimodal tasks~\cite{papadopoulos2023synthetic, chen2022cross}. 
Due to shortcuts between the target and certain modalities, current models often favor trivial solutions~\cite{guo2023modality, chen2022cross}.
While there is no systematic analysis of modality bias in MMBs, prior research has uncovered bias patterns in general multimodal benchmarks like visual question answering (VQA), offering valuable insights into understanding these biases in existing benchmarks.
Two common approaches for analyzing modality bias include automated dataset-level quantification and manual identification by human experts. In automated quantification, \citet{liang2024quantifying} measure modality interaction using information theory and propose two PID estimators to evaluate entire datasets. However, bias can vary significantly across individual samples in MMBs, which limits the ability of dataset-level approaches to detect sample-specific biases. 
Regarding manual identification, 
\citet{goyal2017making} reveal a spurious correlation between text and labels in the VQA~\cite{antol2015vqa} dataset, where simply answering ``yes'' to questions beginning with ``Do you see a ...'' achieves 87\% accuracy without considering the rest of the question or the image. 
Similarly, 
\citet{liu2024tackling} highlight that over 90\% of the answers to questions about whether the audio in the MUSIC-AVQA~\cite{li2022learning} dataset matches the instrument shown in the video are ``yes''. 
\citet{papadopoulos2024verite} simply hypothesize that modality bias in multimodal misinformation benchmarks stems from ``asymmetric pairs'' and they do not make a systematic analysis on the automated bias quantization. Although manual methods can detect and mitigate bias through techniques like data augmentation or filtering rules, they are impractical for analyzing the vast amount of online multimodal misinformation.

Since bias can vary significantly across individual samples, this work investigates the feasibility of automated sample-specific modality bias analysis and makes some interesting observations, providing potential direction and design considerations toward future research.

\subsection{Multimodal Misinformation Benchmarks}
Current multimodal misinformation benchmarks can be broadly categorized into two types: real-world and synthetic datasets. Fakeddit~\cite{nakamura2019r}, the largest multimodal misinformation dataset, contains over 400k samples sourced from the social networking platform Reddit. 
Due to Reddit's aggregation of content from Twitter, Facebook, and WhatsApp, Fakeddit is highly diverse. 
Mocheg~\cite{yao2023end} is a large-scale fact-checking dataset comprising 15,601 claims, each paired with an image and annotated with one of three labels: support, refuted, or not enough information. 
VisualNews~\cite{liu2021visual} is another large-scale dataset, containing over one million news images and captions from four major news agencies: The Guardian, BBC, USA Today, and The Washington Post.
Among synthetic datasets, NewsCLIPings~\cite{luo2021newsclippings} is constructed using techniques such as scene learning and CLIP~\cite{radford2021learning} to produce out-of-context samples. MMFakeBench~\cite{liu2024mmfakebench} leverages powerful vision-language models like DALL-E3~\cite{ramesh2022hierarchical} to generate AI-based misinformation related to textual veracity, visual veracity, and cross-modal consistency distortion. However, as discussed in the introduction, there exists modality bias in these benchmarks, which presents clear drawbacks for training and evaluating MMD models in real-world deployment.

In this paper, we concentrate on the image-text multimodal misinformation dataset and discuss another widely used multimodal type (video-text) in the Appendix~\ref{Discussion}.
We perform the automated analysis on two multimodal misinformation benchmarks: a real-world dataset Fakeddit, and a synthetic dataset MMFakeBench.
By analyzing benchmarks of different scenarios, we seek to comprehensively validate the effectiveness of our automated analysis.
\begin{figure*}[t]
  \centering
  \includegraphics[width=0.9\linewidth]{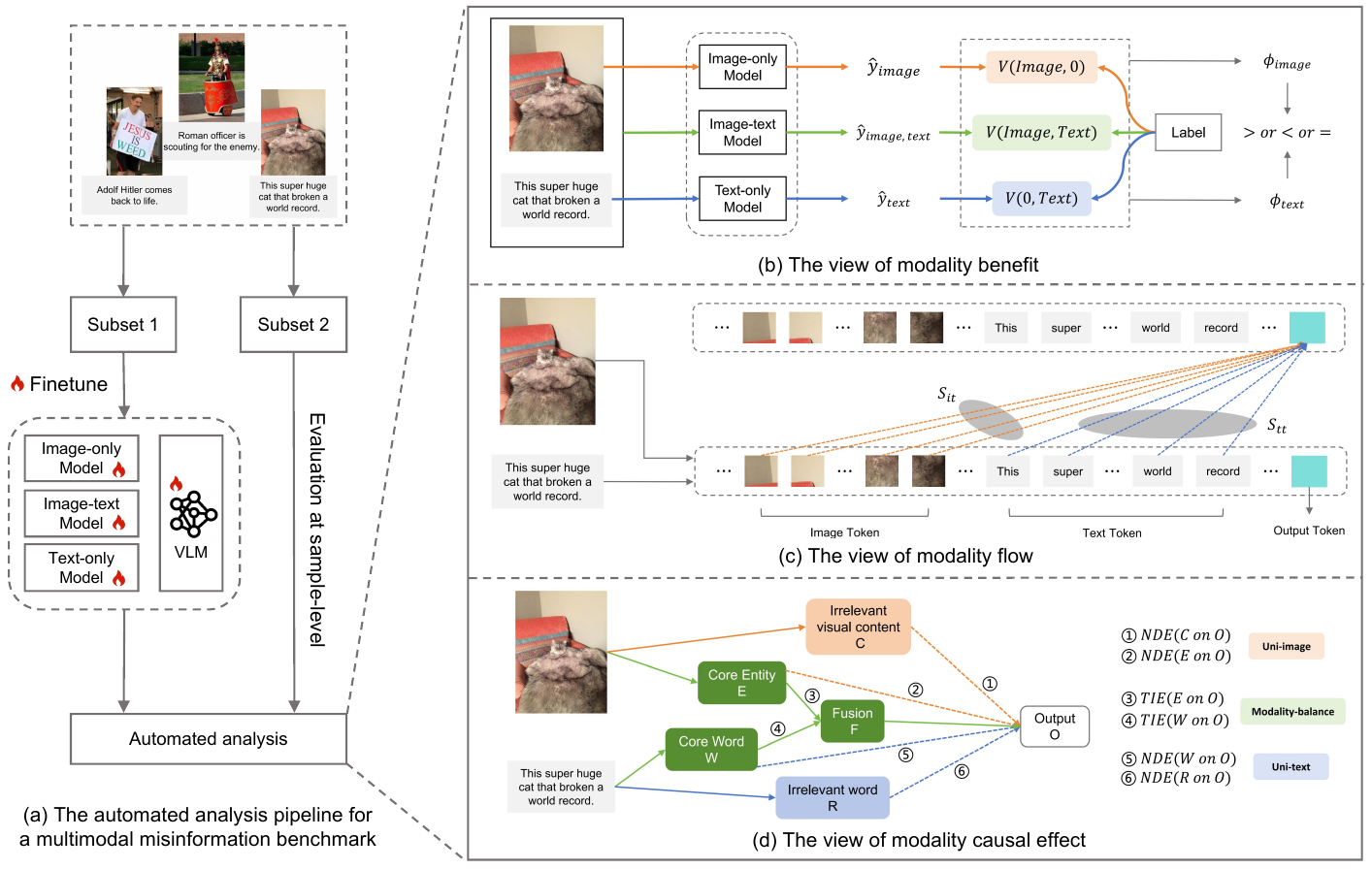}
  \caption{Illustration of proposed automated analysis for modality bias in multimodal misinformation benchmarks.}
  \label{fig: method}
\end{figure*}

\section{Automated Sample-Specific Analysis}

The overall workflow of automated analysis is illustrated in Figure~\ref{fig: method}. Several misinformation detectors are used to power the computation of automated analysis, i.e., the Image-only, Image-text, Text-only, and large vision-language model. We need to fine-tune these models for more reliable measurements because existing models lack robust zero-shot capabilities for MMD.
We randomly select some samples (Subset1) to fine-tune the models and perform single- and multi-view analysis on the remaining subset (Subset2).

\subsection{Modality Benefit}
\label{modalitybenefit}
From the view of modality benefit, we introduce a Shapley value-based metric~\cite{wei2024enhancing, shapley1953value, parcalabescu2023mm}, which is designed for cooperative games, to observe the uni-modal contribution by comparing the model's prediction with/without specific modality.
For generalization, we first illustrate the scenario with $n$ modality and then provide the formula when $n=2$.

Each sample $x=\left( x^{m_1}, x^{m_2}, ..., x^{m_n} \right)$ is with $n$ modality, $y$ is the corresponding label, $x^{m_i}$ is the modality $m_i$ of sample $x$.
Let $M=\{m_1, m_2, ..., m_n\}$ be the set of all modalities, $M'$ be the subset of $M \left( M' \subseteq M\right)$ and $x^{M'}$ be the input sample $x$ with modality set $M'$, we can define a benefit function $V$ that maps the model's prediction with input $M'$ to its benefits:
\begin{align}
V\left ( x^{M'} \right ) = \begin{cases} \left | M' \right | & \text{if} \;   \hat{y} = y, \\0 &\text{otherwise.} \end{cases}
\end{align}
Here $\hat{y}$ is the prediction and $\left | \right|$ denotes the number of $M'$, i.e., if the model makes a correct prediction, the benefit will be the number of input modalities.

Since a player can interact with other players, different permutations of input modalities may yield varying outcomes.
If we define a certain permutation as a strategy and let $\prod_{M}$ be the permutation of $M$, there is $\left | \prod_{M} \right | =n!$ strategies.
For a strategy $\pi \in \prod_{M}$, the marginal benefit of modality $m_i$ of sample $x$ in $\pi$ can be defined as:
$
v\left( \pi; x^{m_i} \right) = V\left ( \pi\left( x^{m_i} \right ) \cup x^{m_i} \right ) - V\left ( \pi\left( x^{m_i} \right ) \right )
$
, where $\pi\left( x^{m_i} \right )$ represents all predecessors of $x^{m_i}$ in $\pi$.
This formula quantifies the increased benefit of modality $x^{m_i}$ compared to its predecessors.
Considering the marginal contribution of modality $m_i$ of sample $x$ in all strategies, the final benefit of modality $m_i$ is given by:

\begin{align}
\phi _{m_i} = \frac{1}{n!} \sum_{\pi \in \prod_{M} } v\left( \pi; x^{m_i} \right)
\end{align}

As shown in Figure~\ref{fig: method}(b), for
image and text modality ($n=2$), there are simply two strategies in $\prod_{M}=\{\pi_1=(m_1, m_2), \pi_2=(m_2, m_1)\}.$
The final contribution of such a specific modality $m_1$ is given by:
\begin{eqnarray}
\begin{aligned}
\phi _{m_1} &=& \frac{1}{2} \left [ v\left( \pi_1; x^{m_1} \right) + v\left( \pi_2; x^{m_1} \right)  \right ] \\
&=& \frac{1}{2} \left [ V\left ( x^{m_1},0^{m_2} \right ) - V\left ( 0^{m_1},0^{m_2} \right )  \right. \\ 
& &\left. + V\left ( x^{m_2},x^{m_1} \right ) - V\left ( x^{m_2},0^{m_1} \right ) \right ]  
\end{aligned}
\end{eqnarray}
where the above $0^{m_i}$ denotes the absence of modality $m_i.$ We adopt zero input for image modality and placeholder padding for text modality following~\citet{wei2024enhancing}. 
We set $V\left ( 0^{image},0^{text} \right )$ to zero and leverage Image-only, Image-text, and Text-only models to compute $V\left ( x^{image},0^{text} \right )$, $V\left ( x^{text},x^{image} \right )$, and $V\left ( x^{text},0^{image} \right )$, respectively.
Finally, we can determine the bias type of each sample, i.e., Uni-image: $\phi _{image} > \phi _{text}$, Modality-balance: $\phi _{image} = \phi _{text}$, Uni-text: $\phi _{image} < \phi _{text}$.


\subsection{Modality Flow}
Figure~\ref{fig: method}(c) depicts the view of modality flow: comparing the information flow from the image/text to the output intuitively reveals whether the model relies more on image or text when making predictions.
Computing accurate attention interaction signals requires advanced models,
so we leverage a large vision-language model (LVLM) rather than smaller models.
From the view of modality flow, we seek to discover the attention interaction pattern between input modalities and target position in a Transformer-based model (Figure~\ref{fig: method}~(c)).
Suppose the input prompt for MMD is $P=\left [..., IT, ..., TT, ..., OT\right ]$, where $IT = \left ( IT_1, IT_2, ..., IT_{n_1}\right )$ is the image token, $TT = \left ( TT_1, TT_2, ..., TT_{n_2}\right )$ is the text token and $OT$ is the output token which is usually the last token.
Comparing the information flow from the image/text to the target token intuitively reveals whether the model relies more on image or text modality when making predictions.
Following \citet{wang2023label}, we employ the saliency score to quantify critical token interactions:
\begin{eqnarray}
S = \left | \sum_{h} A_{h} \odot \frac{\partial \mathcal{L}(P)}{\partial A_{h}}   \right | 
\end{eqnarray}
where $A_h$ represents the attention matrix of $h$-th attention head, $\odot$ is Hadamard product, $P$ is the input prompt, $\mathcal{L}(\cdot)$ is the loss function of multimodal misinformation detection.
Concretely, $S(j_1,j_2)$ denotes the importance of the information flow from $j_2$-th token to $j_1$-th token in the input prompt.
Based on the observation that shallow layers are primarily used for token information aggregation and analysis, and deep layers leverage token information for prediction, we only calculate the saliency score for the last attention layer.
To study the effect of different saliency calculations, we compare our attention-based calculation with a perturbation-based LIME~\cite{ribeiro2016should} in the Appendix~\ref{saliencyCalculation}.

Generally, the number of image tokens exceeds that of text tokens. For instance, a $224 \times 224 $ image can be divided into 64 patch tokens, while the text typically comprises fewer than ten tokens. 
Since most image tokens may represent background information, their individual contribution may be less significant compared to single text tokens. 
To assess the overall contribution, we adopt the sum of the saliency score as the final information flow from the respective modality to prediction.
We also study the effects of different computation strategies
of $S_{it}$ and $S_{tt}$ in the Appendix~\ref{computation_strategies}. 
\begin{align}
S_{it} &= \sum_{k}^{n_1} S(OT,IT_k),\; IT_k \in IT \\
S_{tt} &= \sum_{k}^{n_2} S(OT,TT_k),\; TT_k \in TT 
\end{align}

Then we apply a normalization to $S_{it}$ and $S_{tt}$ to map them to the same interval~\cite{jin2021one}:
\begin{align}
S_{it, norm} = \frac{S_{it}}{S_{it}+S_{tt}},
S_{tt, norm} = \frac{S_{tt}}{S_{it}+S_{tt}}
\end{align}

The value space of saliency scores is continuous, which means $S_{it, norm} \neq S_{tt, norm}$ even when the sample is modality balanced.
Therefore, we define a hyperparameter threshold $\epsilon$ to confine the differences of modality-balanced cases.
In other words, when $\left | S_{it, norm}-S_{tt, norm} \right | < \epsilon $, we consider the sample to be modality-balanced.
We conduct a user study to determine the threshold $\epsilon$ and a detailed description can be found in the Appendix~\ref{user_study}.

\subsection{Modality Causal Effect}
The first two views analyze modality bias from the model's perspective.
Here we introduce a method based on counterfactual reasoning to evaluate modality bias from the reasoning perspective.
The causal mechanisms of MMD problem-solving involve first analyzing the core information, such as primary entities in images and main semantics in text, and then combining them to derive the final prediction. 
However, biased data can yield predictions directly from a single modality.
For instance, in the VQA dataset, irrelevant textual information like ``Do you see a ...'' has a strong connection with the prediction ``yes''. 

In Figure~\ref{fig: method}(d), we illustrate all possible causal reasoning paths for MMD, where different paths correspond to different types of modality bias.
Suppose $I$ is the image, $C$ is the irrelevant visual content of the image, $E$ is the core entity of the image, $T$ is the text, $W$ is the core chunk of the text, $R$ is the irrelevant fragment of the text, $F$ is the information fusion of $E$ and $W$, and $O$ is the output, we make the following definitions.
\textbf{Image Bias:} the model may directly predict through $I \to C \to O$ and $I \to E \to O$.
\textbf{Text Bias:}
the inference paths referred to as text bias include $T \to R \to O$ and $T \to W \to O$.
\textbf{Modality Balance:} the desired causal path is via $I \to E \to F$, $T \to W \to F$ and $F \to O$.
For core information extraction ($C$, $E$, $W$ and $R$), we utilize MiniCPM-V~2.6 and Llama3-8B-Instruct~\cite{llama3modelcard} to process image and text, respectively. 
We provide details of core information extraction in the Appendix~\ref{prompt}.
Then we employ counterfactual reasoning to quantify the causal effects of different paths and identify bias types corresponding to the path exhibiting the greatest causal effect.

Counterfactual reasoning can estimate the causal effect of a treatment variable on a response variable by comparing outcomes under conditions that are different from the factual world.
A detailed explanation of counterfactual reasoning (e.g., the \textit{definition} of counterfactual reasoning and \textit{description of some jargon} that will be used below) can be found in the Appendix~\ref{explanation_counterfactual}.
Here we denote the causal mechanism of MMD as:
$
O_{c, e, w, r, f} = O\left ( C=c, E=e, W=w, R=r, F=f \right ),
f=F_{e, w}=F\left( E=e, W=w \right)
$.

Consider the variable $W$ as an example. There exist two paths between $W$ and $O$, namely $W \to F \to O$ and $W \to O$ in the causal inference graph. Following~\citet{chen2023causal}, we define the total effect (TE) of $W=w$ on $O$ as: 
\begin{align}
TE (\text{W on O}) = O_{w, f} - O_{w^*,f^*}
\end{align}
where $*$ denotes the reference value.
Total Effect can be interpreted as the comparison between two potential outcomes of $W$ under two distinct treatments $w$ and $w^*$.
Meanwhile, Total Effect can be divided into Natural Direct Effect (NDE) and Total Indirect Effect (TIE).
NDE is the causal effect of path $W \to O$, which means information from $W$ to $F$ has been blocked, while TIE denotes the causal effect of path $W \to F \to O$.

In the counterfactual scenario, $W$ is supposed to be values $w$ and $w^*$ simultaneously, where $w^*$ influences the indirect path $W \to F \to O$ and $w$ influences the direct path $W \to O$. In other words, $w^*$ isolates the influence of $W$ on the intermediate factor $F$, thus observing the effect of $W$ on $O$:
\begin{align}
NDE (\text{W on O}) &= O_{w,f^*} - O_{w^*,f^*}\\
TIE (\text{W on O}) &= TE(\text{W on O}) - NDE(\text{W on O}) \nonumber \\
&= O_{w,f} - O_{w,f^*}
\end{align}

Following previous studies~\cite{chen2023causal, wang2021clicks}, we set other variables $C$, $E$, and $R$ to their reference value $c^*$, $e^*$, and $r^*$ when $W=w^*$. For reference value, we adopt zero input for $c^*$ and $e^*$, and placeholder padding for $w^*$ and $r^*$.
To obtain the output prediction, we apply a non-linear fusion strategy. For example,$
O_{c,e,w,r,f} = \mathcal{F} \left ( O_c, O_e, O_w, O_r, O_f \right )
= tanh(O_c) + tanh(O_e) + O_f + tanh(O_w) + tanh(O_r)
$, where $\mathcal{F}(\cdot)$ is the non-linear fusion strategy, $O_c$ is the output logit of the irrelevant visual context branch, $O_e$ is the outcome of the core entity branch, $O_w$ is the result of the core semantic words branch, $O_r$ is the output of the irrelevant word branch, $O_f$ is the output of fusion branch.
To compute these outputs, we utilize the Image-only model for $O_c$ and $O_e$, the Text-only model for $O_w$ and $O_r$, and the Image-text model for $O_f$. While $\mathcal{F}(\cdot)$ can be any differentiable binary function, \citet{chen2023causal} observe that tanh-sum yields the best performance.

Similarly, we can compute the natural direct effect of variable $C$, $E$, and $R$ on $O$ and the total indirect effect of variable $E$ on $O$, i.e., $NDE (\text{C on O})$, $NDE (\text{E on O})$, $NDE (\text{R on O})$, and $TIE (\text{E on O})$.
Figure~\ref{fig: method}(d) shows that these causal effect items correspond to the six distinct paths within the inference graph, with each path associated with a specific modality bias type. For each sample, we determine the bias type based on the path exhibiting the greatest causal effect.
Finally, multi-view analysis is derived through a prior majority voting, where the outcome is determined by the majority of views. 
In the tie event, priority is assigned to the category with the larger number of samples.
Discussion of more ensemble strategies is shown in the Appendix~\ref{ensemble}.

\section{Experiment Setting}
\subsection{Benchmarks}
We conduct the automated sample-specific modality bias analysis on two MMBs, i.e., Fakeddit and MMFakeBench. Fakeddit is a real-world benchmark and contains over six hundred thousand multimodal samples. Moreover, MMFakeBench is a synthetic dataset generated by large vision-language models like DALL-E3. These two benchmarks are particularly representative due to their large scale (~680K samples) and extensive coverage of diverse domains, including real-world misinformation, AI-generated synthetic content, satire, rumors, face swaps, and Photoshop-edited images. A detailed description of these datasets, along with their statistical distributions, is provided in the Appendix~\ref{Statistics}.




\subsection{Models}
We define the required misinformation detection models for automated analysis as \{Image-only, Image-text, Text-only, LVLM\}. For efficiency, we use the first three types of models to support the view of modality benefit and causal effect~\cite{niu2020counterfactual}. As for modality flow, computing accurate attention interactions requires advanced models, so we leverage a large vision-language model (LVLM) rather than smaller models. We select the following models for experiments, i.e., \textbf{Image-only}: UnivFD~\cite{ojha2023towards} and DT(I); \textbf{Image-text}: HAMMER~\cite{shao2023detecting} and DT(I, T)~\cite{papadopoulos2024verite}; \textbf{Text-only}: FFNews~\cite{huang2022faking} and DT(T); \textbf{LVLM}: MiniCPM-V~2.6~\cite{yao2024minicpm}. Since existing models possess limited zero-shot performance, we first fine-tune these models for reliability. We describe details of selected models, the selection criteria, and the fine-tuning process in the Appendix~\ref{finetune_detail}.



\begin{table*}[h]
\centering
\caption{The quantification comparison of automated analysis and human judgment. We report the predicted proportion (without []) and accuracy (within []) of different bias types compared to human annotations.
The proportion ranges from 0 to 1, accuracy and F1 are presented as percentages (\%). UI, MB, and UT represent Uni-image, Modality-balance, and Uni-text, respectively.}
\setlength{\tabcolsep}{0.8mm}
\begin{tabular}{@{}c|ccccc|ccccc@{}}
\toprule
       & \multicolumn{5}{c|}{Fakeddit}                 & \multicolumn{5}{c}{MMFakeBench}               \\ \midrule
       & UI & MB & UT  & Acc & F1 & UI & MB & UT & Acc & F1 \\ \midrule
Human          &0.15&0.75& 0.10&-&-    &0.13&0.74&0.13&-&-       \\
Modality benefit &0.04[0.00]&0.85[78.43]&0.11[44.12]&71.67& 43.40&0.42[11.02]&0.42[80.95]&0.16[63.83]&48.67&47.98   \\
Modality flow &0.13[42.50]&0.52[96.15]&0.35[28.85]&65.67& 54.51&0.04[0.00]&0.67[70.00]&0.29[12.50]&50.33&27.85   \\
Modality causal effect &0.33[29.29]&0.61[87.43]&0.06[27.78]&64.67&46.51&0.08[37.50]&0.65[82.47]&0.27[43.90]&68.33&54.89   \\
Multi-view analysis &0.10[56.67]&0.78[85.53]&0.12[57.14]&\textbf{79.33}&\textbf{64.75}&0.04[69.23]&0.80[85.77]&0.16[75.00]&\textbf{83.33}&\textbf{68.44}   \\
\midrule
Benefit-Flow   &0.04[0.00]&0.88[79.09]&0.08[57.69]&74.33&46.27&0.14[0.00]&0.84[72.91]&0.02[0.00]&61.00&25.85    \\
Benefit-Causal &0.05[0.00]&0.92[76.09]&0.03[0.00]&70.00&27.94&0.15[20.45]&0.71[84.11]&0.14[71.43]&73.00&59.00    \\
Flow-Causal    &0.22[35.38]&0.72[86.64]&0.06[27.78]&72.00&49.24&0.00[0.00]&0.96[74.56]&0.04[46.15]&73.33&35.49    \\
\bottomrule
\end{tabular}
\label{tab:result}

\end{table*}

\subsection{Implement Details}

We conduct our experiments with the following models: \{UnivFD, HAMMER, FFNews, MiniCPM-V~2.6\}.
All experiments are conducted on one A800 80GB GPU. 
The inference time of modality benefit, flow, and causal effect: 1 hour, 3 hours, and 2 hours every 60k samples.
Our framework is adaptable to any misinformation detection method and LVLM.
Especially for LVLM, it can be easily generalized to language models by applying different prompts to flexibly function them as Image-only, Image-text, and Text-only models. 
In the Appendix~\ref{extend_to_large_models}, we further investigate the performance of pure language model-driven analysis.

\subsection{Evaluation}
We are the first to propose an automated sample-specific modality bias analysis and no existing baselines are available for direct comparison. 
The approach most similar to ours is presented in~\cite{liang2024quantifying}, which is specifically designed to quantify bias across the entire dataset and rely on computations based on the joint distribution. This method is unsuitable for sample level. 
Therefore, we conduct a human evaluation with three annotators to validate the alignment of single- and multi-view analysis and human judgment. 
To assess the reliability and agreement of human annotations, we conducted Krippendorff's alpha test~\cite{krippendorff2011computing}. 
Details of annotators' demographic characteristics, annotation procedure, and the result of Krippendorff's alpha test can be found in the Appendix~\ref{human_evaluation}.
Following~\cite{liang2024quantifying}, we randomly select 300 samples from each dataset to conduct the experiment.
We report the predicted proportions of each modality bias type and the percentage that aligns with human judgment.
For example, $0.78[85.53]$ denotes that multi-view analysis classifies 0.78 of the samples as modality-balance, and among these samples, 85.53\% of the results are consistent with human judgment. We also report the overall accuracy and F1 score.

\section{Experimental Results}
This section contains three interesting findings about our proposed automated sample-specific modality bias analysis. More ablation experiments (i.e., saliency score calculations, and computation strategies of $S_{it}$ and $S_{tt}$ in modality flow, the effect of ensemble strategies), an error analysis and a post-validation can be found in the Appendix~\ref{more_ablation_study},~\ref{error_analysis},~\ref{multi_detectors}, which helps further understand the underlying mechanisms of our automated sample-specific bias analysis.

\subsection{Key to Reliable Automated Analysis}
\label{key}
Table~\ref{tab:result} depicts the quantification comparison of automated analysis and human judgment.

\textbf{Comparison of Proportion.}
According to human judgment, most samples are modality-balanced, while only a small proportion are biased. Although single-view analysis generally follows this pattern, notable differences exist in specific numerical values. For example, on Fakeddit, modality benefit classifies 0.04 of the samples as ``Uni-image'', modality flow classifies 0.35 of the samples as ``Uni-text'', and modality causal effect classifies 0.33 of the samples as ``Uni-image''. A similar trend is observed on MMFakeBench. However, multi-view analysis integrates the strengths of each individual view, yielding results that most closely align with human judgment.

\textbf{Comparison of Accuracy.} Different views reveal distinct patterns of bias, and single-view analysis may underperform in certain scenarios. For example, the modality benefit analysis shows strong performance (71.67\%) on Fakeddit while weak performance (48.67\%) on MMFakeBench. However, the ensemble multi-view analysis consistently achieves the highest performance across both datasets, underscoring the stability of multi-view approaches in the complex task of automatically detecting modality bias across diverse scenarios, including both real-world and synthetic samples.

\textbf{Ablation Study.} We also conduct an ablation study to assess the contribution of each view: (1) Benefit-Flow: Omitting the modality causal effect. (2) Benefit-Causal: Removing the modality flow. (3) Flow-Causal: Excluding the modality benefit. As shown at the bottom of Table~\ref{tab:result}, each view contributes meaningfully to the multi-view analysis. 

Multi-view analysis significantly outperforms the three single-view methods in both performance and stability. Therefore, we conclude that automated sample-specific modality bias analysis is a complex task for machines. While reliable measurements cannot be attained solely through single-view analysis, ensemble multi-view demonstrates promising potential for real-world deployment.



\begin{table}[]
\centering
\caption{The accuracy [\%] of modality benefit, modality causal effect, and multi-view analysis under different choices of misinformation detectors. G$i$ denotes Group$i$.}
\begin{tabular}{@{}c|ccccc@{}}
\toprule
Fakeddit & G1 & G2 & G3 & G4 \\ \midrule
Modality benefit  & 71.67 & 64.67 & 68.67 & 54.67\\
Modality causal effect & 64.67 & 60.33 & 61.67 & 58.67\\
Multi-view analysis & 79.33 & 68.33 & 76.67 & 74.67\\
\midrule
MMFakeBench & G1 & G2 & G3 & G4\\ 
\midrule
Modality benefit  & 48.67 & 40.67 & 48.67 & 61.67\\
Modality causal effect& 68.33 & 70.33 & 44.00 & 62.33\\
Multi-view analysis & 83.33 & 71.67 & 60.00 & 63.00\\
\bottomrule
\end{tabular}
\label{tab:robustness}
\end{table}

\subsection{Vulnerability to Detector Fluctuations}
\label{vulnerability}
During the computation of automated analysis, various misinformation detectors are involved, e.g., the image-only, image-text, and text-only models of modality benefit and modality causal effect, as well as the LVLM of modality flow. A pertinent question arises: \textbf{is automated analysis robust to different choices of misinformation detectors?}

To answer this question, we evaluate the sensitivity of \emph{modality benefit}, \emph{modality causal effect}, and \emph{multi-view analysis} by altering specific models and observing the change in accuracy.
We select four model combinations (across Image-only, Image-text, and Text-only models): 
\begin{itemize}
    \item Group1=\{UnivFD, HAMMER, FFNews\}
    \item Group2=\{\textbf{DT(I)}, HAMMER, FFNews\}
    \item Group3=\{UnivFD, \textbf{DT(I, T)}, FFNews\}
    \item Group4=\{UnivFD, HAMMER, \textbf{DT(T)}\}
\end{itemize}

As illustrated in Table~\ref{tab:robustness}, the maximum fluctuation of performance on Fakeddit and MMFakeBench exceeds 10\% for both single-view and multi-view analysis, indicating that automated analysis is prone to detector-induced fluctuations. We take this phenomenon as unavoidable because each view quantifies modality bias based on models' output, and the performance of different models can vary significantly. Transferring the model for a specific modality inevitably affects the distribution of prediction for that modality, which in turn influences the calculation of modality contribution.

Therefore, in practical applications, certain improvements are necessary to enhance the robustness of automated analysis. On the one hand, the simplest approach is to ensemble various misinformation detectors for each view, thus leveraging the strengths of different types of detectors. 
In the Appendix~\ref{multi_detectors}, we empirically investigate the impact of ensembling various detectors per model category, with experimental results demonstrating enhancements in the robustness of our automated analysis.
However, this method introduces additional computational overhead and is more suitable for scenarios where real-time consideration is low-priority, such as preliminary cleaning of modality-biased benchmarks. On the other hand, model-agnostic features can be incorporated to compute detectors' output, such as edge or texture features for images and TF-IDF features for text. While this reduces reliance on specific model architectures, it requires the design of effective model-agnostic feature extraction methods to ensure that these features can capture the key information related to modality bias.

\subsection{Modality-balanced  vs. Biased Samples}
\label{divergence}

Table~\ref{tab:result} reveals that multi-view analysis achieves high accuracy on modality-balanced samples but exhibits relatively lower accuracy on biased ones. For example, on Fakeddit, the accuracy of multi-view analysis on ``Modality-balance'' samples is 85.53\%, while on ``Uni-text'' samples, the accuracy drops to 57.14\%. A similar trend is observed on MMFakeBench, where the accuracy on ``Modality-balance'' samples is 85.77\%, but on ``Uni-image'' samples, it decreases to 69.23\%. \textbf{What contributes to this discrepancy?}

\begin{figure}[t]
  \centering
  \includegraphics[width=\linewidth]{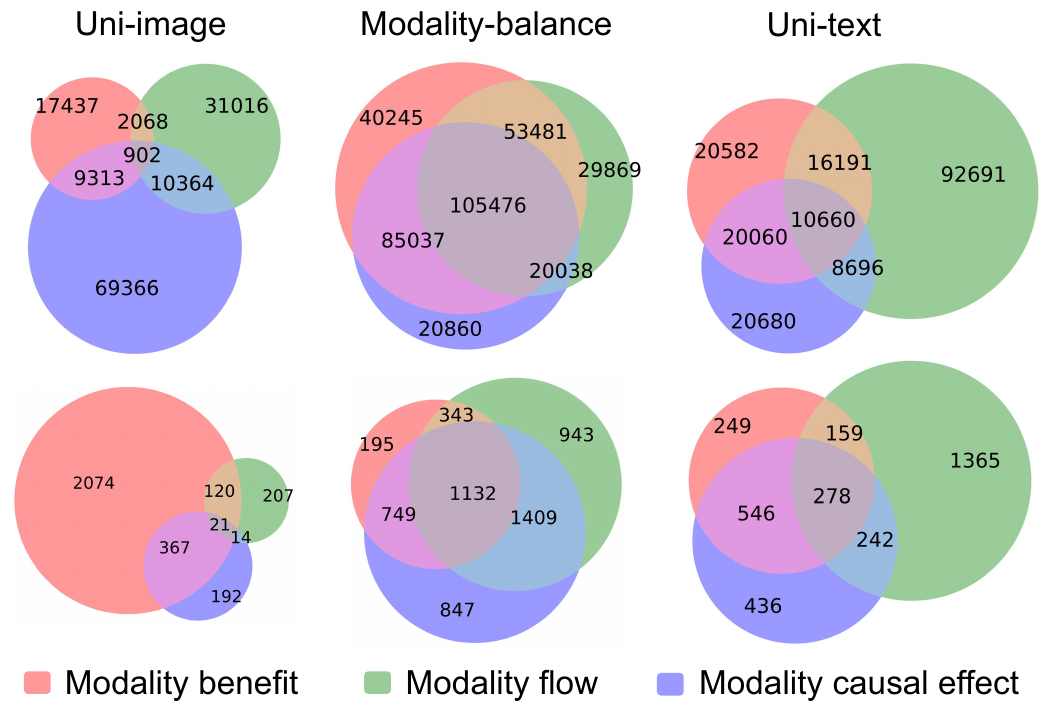}
  \caption{The Venn diagram of three single-views on Fakeddit (top three) and MMFakeBench (bottom three).}
  \label{fig: consistency}
\end{figure}

To answer this question, we use Venn diagrams to visualize the intersections among different views to analyze the consistency of multi-view analysis. It is important to note that this analysis encompasses the entire dataset, rather than those samples from human evaluations. As illustrated in Figure~\ref{fig: consistency}, different views exhibit high alignment on modality-balanced samples but significant divergence on biased samples. We attribute this divergence to the fact that different views possess distinct patterns for capturing bias. Generally, higher consistency among views yields higher accuracy, and thus, this divergence leads to suboptimal accuracy on biased samples. In real-world deployment, if our objective is to clean a modality-biased benchmark by retaining only modality-balanced samples, the results of the automated analysis can serve as a robust reference. Conversely, if the focus is on biased samples, it becomes necessary to employ related techniques to mitigate this divergence, thereby ensuring the reliability of the results. For instance, a calibrator could be designed to post-process the predicted probabilities of biased samples of each view.

\section{Conclusion}
In this work, we investigate whether it is possible to establish an automated sample-specific modality bias analysis for existing multimodal misinformation benchmarks.
We first propose three quantification methods based on different theories and adapt them to bias identification, i.e., the view of modality benefit, modality flow, and modality causal effect.
Then we conduct a human evaluation on two multimodal misinformation benchmarks to study the practicability of automated analysis and derive three interesting findings that offer design considerations and improvement direction toward future research.
Experimental results indicate that automated sample-specific modality bias analysis holds potential for practical applications. 
This suggests its capability to perform tasks like dataset cleaning (e.g., retaining modality-balanced samples) to mitigate the severity of modality bias.

\section{Limitations}
There is one primary limitation of this work. We do not study the effect of different large vision-language models (e.g., larger and stronger LVLMs) on the view of modality flow because of LVLMs' high computation cost of saliency score calculation based on the loss backward process, which we encourage future work to investigate.
{
    \small
    \bibliographystyle{ieeenat_fullname}
    \bibliography{main}
}

\clearpage
\setcounter{page}{1}
\maketitlesupplementary

\section{Description of Appendix}
\label{Guide}
This appendix contains the discussion of some considerations, the investigation of different settings, the detailed information about corresponding processes, the extension to pure large models-driven architecture, the error analysis of multi-view output, the post-validation, and the multi-detectors ensemble, which contributes to a comprehensive understanding of this paper. 

Appendix~\ref{Discussion} discusses several considerations of this work, such as the definition of modality bias, the multimodal video misinformation benchmarks, and the versatility of our proposed automated analysis.
Appendix~\ref{saliencyCalculation} delves into the effect of different saliency score calculation methods. 
Appendix~\ref{computation_strategies} studies the effect of different computation strategies of $S_{it}$ and $S_{tt}$ in the view of modality flow.
Appendix~\ref{ensemble} examines how various methods of combining multi-view can influence performance. 
Appendix~\ref{user_study} describes the determination and impact of super-hyperparameter $\epsilon$.
Appendix~\ref{prompt} focuses on the core information extraction prompts and the effect of different extraction model combinations.
Appendix~\ref{explanation_counterfactual} explains the definition of counterfactual reasoning and some jargon used in our computation process. 
Appendix~\ref{Statistics} provides a quantitative overview of multimodal misinformation benchmarks utilized in our work. 
Appendix~\ref{finetune_detail} detailedly clarifies the model description, model selection criteria, and fine-tuning details.
Appendix~\ref{human_evaluation} presents the details of human annotation and instruction.
Appendix~\ref{extend_to_large_models} extends our automated analysis to pure large models-driven architecture and discusses the performance.
Appendix~\ref{error_analysis} conducts an error analysis of the ensemble multi-view analysis.
Appendix~\ref{post_validation} provides an alignment test between our automated analysis and the behaviors of misinformation detector models.
Appendix~\ref{multi_detectors} ensembles multiple misinformation detectors for each model category to enhance the robustness of our automated analysis.

\section{Discussion}
\label{Discussion}
Firstly, the definition of ``modality bias'' is derived from~\cite{guo2023modality}, referring to the tendency of a model to rely on a single modality (e.g., image or text) for decision-making. However, there might be multiple forms of modality bias in practical applications according to varying definitions. Theoretically, each view (i.e., Modality benefit, Modality flow, and Modality causal effect) holds a distinct bias recognition pattern, so the ensemble multi-view analysis is robust to diverse forms of bias.

Secondly, while MM-SHAP explores a Shapley value-based metric for measuring multimodal contributions in vision and language models, the effectiveness of Shapley value in multimodal misinformation benchmarks and whether relying solely on Shapley value sufficiently captures biases remain underexplored. Our work provides a systematic analysis of such automated modality bias quantization for multimodal misinformation benchmarks.

Thirdly, there is another widely used multimodal misinformation type: video and text modality, e.g., FakeSV~\cite{qi2023fakesv} and FakeTT~\cite{bu2024fakingrecipe}.
We do not conduct experiments on the multimodal video dataset for two primary reasons:
(1) Most multimodal video misinformation detection models typically involve a combination of multiple branches, such as text and video. The performance of using only the video branch tends to be relatively poor; for instance, the video-only model achieves an accuracy of only 69.05\% on FakeSV~\cite{qi2023fakesv}. However, to accurately assess the contribution of the video modality, we need to utilize the video-only model independently. Without a reliable accuracy performance, the bias analysis would also be deemed unreliable.
(2) Most existing video misinformation samples introduce misinformation through modifications of key frames. Consequently, misinformation detection methods that rely on key frames can achieve satisfactory performance in misinformation detection~\cite{bu2024fakingrecipe}. Distinguishing bias types between key frames and text, which can be solved by our image-text scenario analysis, is meaningful for these multimodal video misinformation datasets.
As this work is the first exploration into whether automated analysis can yield reasonable measurements, we concentrate on the most fundamental and widely applicable scenario: the image-text setting. Transitioning to a video-text scenario necessitates a more refined design, which we encourage future research to investigate.

Fourthly, from the view of modality benefit, we can determine the type of modality bias by comparing the final output benefit of image modality and text modality. Nevertheless, when $V\left ( x^{m_1},0^{m_2} \right ), V\left ( 0^{m_1},0^{m_2} \right ), V\left ( x^{m_2},x^{m_1} \right ), $ and $ V\left ( x^{m_2},0^{m_1} \right )$ all equal zero, the model is unable to make accurate predictions. In such cases, we hypothesize that the difficulty of samples exceeds the discriminative capacity of models, and the Shapley value cannot provide a reasonable classification. 

Fifthly, we investigate the automated sample-specific modality bias analysis for multimodal misinformation benchmarks. This deepens our understanding of such benchmarks and provides new insights for online multimodal content analysis. However, this method can be applied not only in the field of misinformation detection. Our automated analysis is broadly applicable to general tasks like visual question answering (VQA) and extends to other modalities like audio.

Sixthly, while our work focuses on identifying and analyzing modality bias, improving misinformation detection based on bias analysis is a direction worthy of in-depth exploration. We encourage future work to improve model training by leveraging modality bias analysis results as auxiliary labels during the optimization process of multimodal misinformation detection.

Seventhly, in real-time applications, the primary computation cost arises from the inference of large models. While the forward of modality flow involves a MiniCPM-V 2.6, the modality causal effect incorporates both MiniCPM-V 2.6 and Llama3-8B-Instruct. This results in a relatively slower inference speed for these two views. A potential approach is to utilize quantized versions of large models in real-time applications to reduce computational costs. 
In fact, our primary application scenarios do not prioritize real-time performance. Researchers can use automated analysis to clean biased benchmarks by retaining modality-balanced samples solely. Because this is a one-time cleaning, it is acceptable with a not fast inference speed.
Additionally, the speed of automated analysis is sufficiently efficient when compared to traditional methods of manually summarizing bias patterns and identifying biased samples at the sample level.

\section{More Ablation Study}
\label{more_ablation_study}
\subsection{Effect of Saliency Score Calculations}
\label{saliencyCalculation}
Table~\ref{tab:saliencyCalculation} presents the results of our saliency score calculations and LIME for comparative analysis, specifically focusing on multi-view analysis and inference speed. FPS (Frame Per Second) denotes the number of samples that can be processed per second (i.e., a higher value indicates faster). The choice of saliency score calculation method has relatively little impact on the inference speed compared to the performance of multi-view analysis.

\begin{table*}[t]
\centering
\caption{The effect of different saliency score calculations on the multi-view analysis. UI, MB, and UT represent Uni-image, Modality-balance, and Uni-text, respectively.}
\setlength{\tabcolsep}{1.3mm}
\begin{tabular}{@{}c|ccccc|ccccc|c@{}}
\toprule
       & \multicolumn{5}{c|}{Fakeddit}                 & \multicolumn{5}{c|}{MMFakeBench} & Speed              \\  \midrule
 & UI & MB & UT  & Acc & F1 & UI & MB & UT & Acc & F1 & FPS\\ \midrule
Ours & 0.10[56.67]&0.78[85.53]&0.12[57.14]&\textbf{79.33}&\textbf{64.75}&0.04[69.23]&0.80[85.77]&0.16[75.00]&\textbf{83.33}&\textbf{68.44} & \textbf{0.4942} \\
LIME & 0.05[60.00] & 0.92[76.09] & 0.03[0.00] & 73.00& 37.94 & 0.04[69.23] & 0.82[83.67] & 0.14[71.43] & 81.33 & 65.19 &0.3489  \\
\bottomrule
\end{tabular}
\label{tab:saliencyCalculation}
\end{table*}

\subsection{Effect of Computation Strategies}
\label{computation_strategies}
As for the computation strategies of $S_{it}$ and $S_{tt}$, we report the predicted proportion under sum, average and maximum conditions in Table~\ref{tab:computation_stratgies}.
We observe that the results of average and maximum strategies are highly unreasonable, which exhibits a strong bias toward text modality. 
We refer to this phenomenon as the modality gap. 
For instance, the image modality typically contains more tokens than the text modality, but many of these tokens often carry background information with minimal impact on the output. 
When using the average strategy, the contribution of the text modality is exaggerated.
A similar problem arises with the maximum strategy, likely due to inherent differences in how the LVLM assigns attention to individual tokens of different modalities. 
This could be attributed to the fact that LVLMs consist of a superior language model (>7B) paired with a simple small image encoder (500M). 

\begin{table}[]
\centering
\caption{The predicted proportion [0-1] of modality flow under different aggregation computation strategies.}
\setlength{\tabcolsep}{1.8mm}
\begin{tabular}{@{}c|ccc@{}}
\toprule
Fakeddit &Uni-image&Modality-balance&Uni-text \\ \midrule
Sum(Ours) & 0.13  & 0.52  & 0.35  \\
Avg       & 0.00  & 0.01  & 0.99  \\
Max       & 0.09  & 0.23  & 0.68  \\ \midrule
MMFakeBench &Uni-image&Modality-balance&Uni-text \\ \midrule
Sum(Ours) & 0.04  & 0.67  & 0.29  \\
Avg       & 0.00  & 0.00  & 1.00  \\
Max       & 0.03  & 0.33  & 0.64  \\ 
\bottomrule
\end{tabular}
\label{tab:computation_stratgies}
\end{table}

\subsection{Effect of Ensemble Strategies}
\label{ensemble}

We explore the impact of different ensemble strategies in Table~\ref{tab:ensembleResult}, including random majority voting, prior majority voting (ours), and weighted voting. The weights assigned to each view are [0.3, 0.2, 0.5], which are determined based on the average performance of single-view analysis. For instance, modality causal effect ranks second on Fakeddit and first on MMFakeBench, demonstrating overall superior performance among the three single-view analyses. Therefore, we assign a weight of 0.5 to this view. Different voting strategies exhibit varying performance across different benchmarks. Overall, prior majority voting demonstrates the most stability and optimal performance.

\begin{table*}[t]
\centering
\caption{The effect of different ensemble strategies on the multi-view analysis. UI, MB, and UT represent Uni-image, Modality-balance, and Uni-text, respectively.}
\setlength{\tabcolsep}{0.41mm}
\begin{tabular}{@{}c|ccccc|ccccc@{}}
\toprule
       & \multicolumn{5}{c|}{Fakeddit}                 & \multicolumn{5}{c}{MMFakeBench}               \\ \midrule
Ensemble Strategy & UI & MB & UT  & Acc&F1 & UI & MB & UT & Acc&F1 \\ \midrule
Random majority voting & 0.14[44.19]&0.70[90.00]&0.16[51.06]&77.33&63.49 & 0.11[26.47]&0.66[84.26]&0.23[53.62]&70.67&57.68\\
Prior majority voting (Ours) & 0.10[56.67]&0.78[85.53]&0.12[57.14]&\textbf{79.33}&\textbf{64.75}&0.04[69.23]&0.80[85.77]&0.16[75.00]&\textbf{83.33}&\textbf{68.44}   \\
Weighted voting & 0.20[37.70]&0.68[91.18]&0.12[57.14]&76.33 &63.88& 0.05[69.23]&0.71[84.11]&0.24[49.32]&75.00&60.40   \\
\bottomrule
\end{tabular}
\label{tab:ensembleResult}
\end{table*}

\section{Detailed Methodology}
\subsection{Determination of Threshold}
\label{user_study}

\begin{figure}[t]
  \centering
  \includegraphics[width=\linewidth]{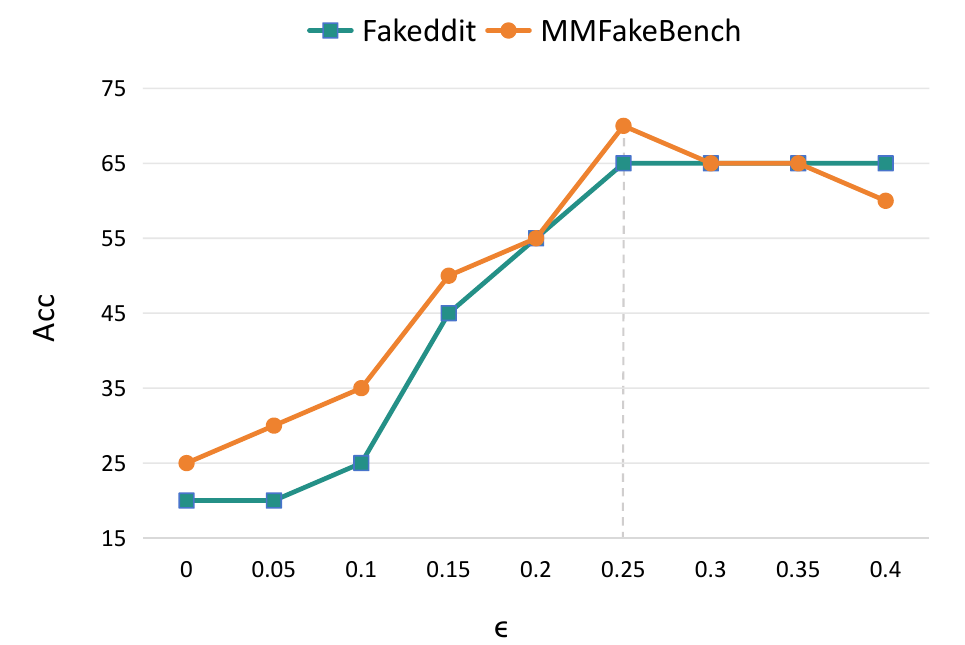}
  \caption{Accuracy of the view of modality flow with varying threshold $\epsilon$ on Fakeddit and MMFakeBench.}
  \label{fig: threshold}
\end{figure}

We conduct a user study to determine the threshold in the view of modality flow, selecting 20 samples from Fakeddit and MMFakeBench and manually annotating the types of modality bias. 
It is important to note that these samples are used for tuning the threshold and are different from those used for human evaluation.
In this user study, the first author of this paper serves as the data annotator and adopts the same criteria described in Appendix~\ref{human_evaluation}.
By adjusting the threshold from 0 to 0.4 in increments of 0.05, we identify the threshold that achieves the highest accuracy for the modality flow analysis.
As shown in Figure~\ref{fig: threshold}, we set the threshold as 0.25.
We also present the results of the ensemble multi-view analysis under different threshold $\epsilon$ in Table~\ref{tab:thresholdsOnResults}. The general trend observed is that, as the threshold increases, accuracy initially rises, then stabilizes, and eventually declines. It is consistent with the findings from the above user study (Figure~\ref{fig: threshold}).

\begin{figure}[t]
  \centering
  \includegraphics[width=\linewidth]{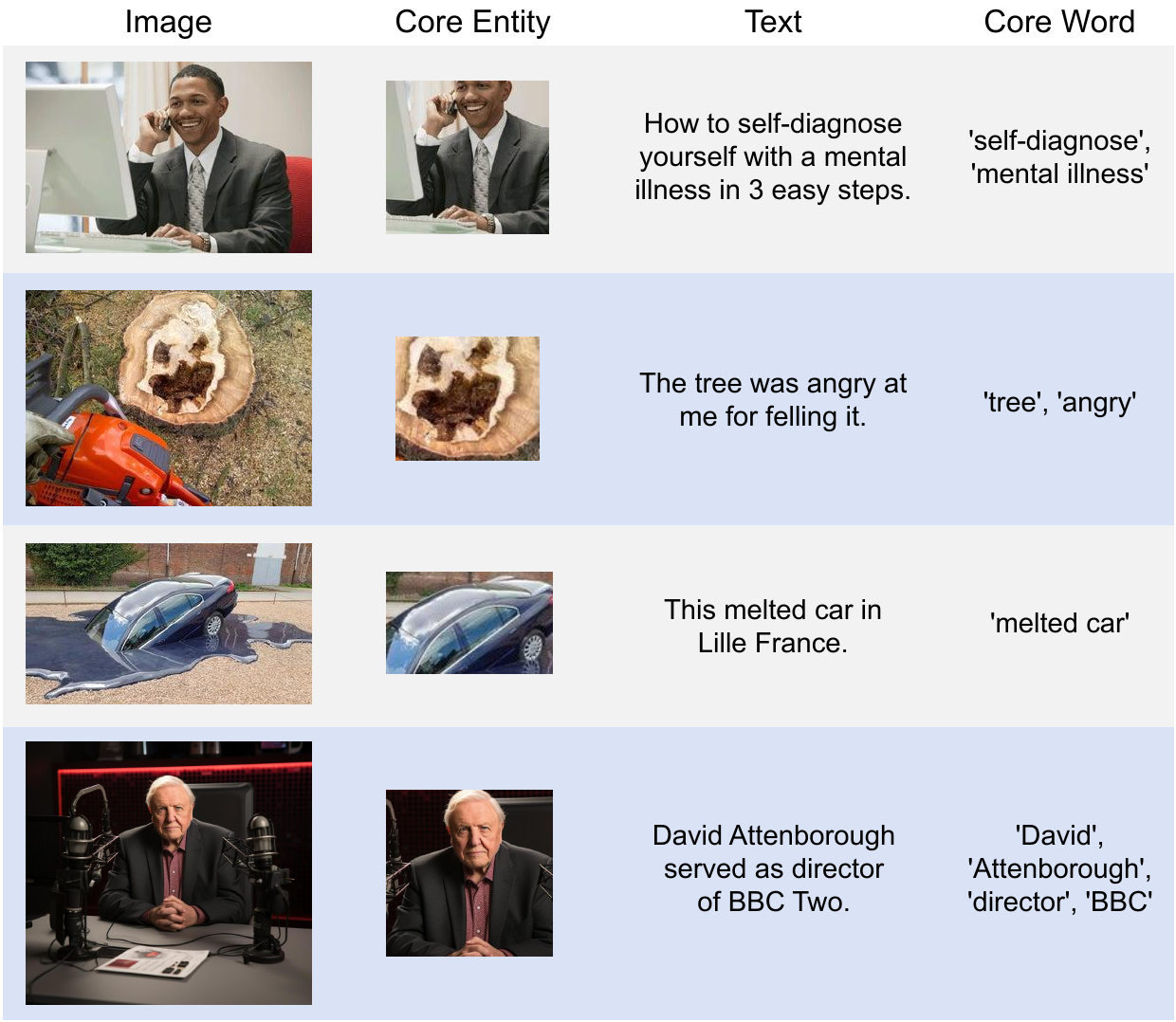}
  \caption{Examples of core information extraction.}
  \label{fig: exampleOfCore}
\end{figure}

\begin{table*}[h!]
\centering
\caption{The effect of different threshold $\epsilon$ on the multi-view analysis. UI, MB, and UT represent Uni-image, Modality-balance, and Uni-text, respectively.}
\setlength{\tabcolsep}{1.55mm}
\begin{tabular}{@{}c|ccccc|ccccc@{}}
\toprule
       & \multicolumn{5}{c|}{Fakeddit}                 & \multicolumn{5}{c}{MMFakeBench}               \\ \midrule
$\epsilon$ & UI & MB & UT  & Acc&F1 & UI & MB & UT & Acc&F1 \\ \midrule
0 & 0.14[40.48]&0.72[84.19]&0.14[46.51]&72.67&58.72 & 0.21[22.58]&0.56[82.74]&0.23[51.43]&63.00&54.78 \\
0.05 & 0.14[40.48]&0.72[84.19]&0.14[46.51]&72.67&58.72 & 0.15[20.45]&0.62[81.72]&0.23[51.43]&65.67 &53.84  \\
0.10 & 0.13[44.74]&0.75[85.02]&0.12[57.14]&76.67&62.63 & 0.15[20.45]&0.62[81.72]&0.23[51.43]&65.67&53.84   \\
0.15 & 0.10[56.67]&0.78[85.53]&0.12[57.14]&\textbf{79.33}&\textbf{64.75} & 0.11[27.27]&0.71[84.11]&0.18[67.92]&75.00&61.57   \\
0.20 & 0.10[56.67]&0.78[85.53]&0.12[57.14]&\textbf{79.33}&\textbf{64.75} & 0.08[37.50]&0.76[85.09]&0.16[75.00]&79.67&65.45   \\
0.25 (Ours) &0.10[56.67]&0.78[85.53]&0.12[57.14]&\textbf{79.33}&\textbf{64.75}&0.04[69.23]&0.80[85.77]&0.16[75.00]&\textbf{83.33}&\textbf{68.44}   \\
0.30 & 0.10[56.67]&0.78[85.53]&0.12[57.14]&\textbf{79.33}&\textbf{64.75} &0.04[69.23]&0.80[85.77]&0.16[75.00]&\textbf{83.33}&\textbf{68.44}   \\
0.35 & 0.07[40.91]&0.82[82.86]&0.11[60.61]&77.33&58.91 & 0.04[69.23]&0.80[85.77]&0.16[75.00]&\textbf{83.33}&\textbf{68.44}    \\
0.40 & 0.07[40.91]&0.84[81.20]&0.09[53.57]&75.67&54.69 & 0.04[69.23]&0.82[83.67]&0.14[71.43]&81.33&65.19    \\
\bottomrule
\end{tabular}
\label{tab:thresholdsOnResults}
\end{table*}

Additionally, we explore an automated clustering method, K-Means, as an alternative to the threshold ($\epsilon$) for comparison. 
Firstly, as indicated in Table~\ref{tab:kmeans}, the performance of k-means is significantly inferior to that of the threshold ($\epsilon$). Secondly, the process of manually setting the threshold ($\epsilon$) is straightforward and entails a small workload. Once established, the analysis is fully automated during the inference phase. Considering these two points, we believe that the manual setting of the threshold ($\epsilon$) is reasonable.

\begin{table*}[h!]
\centering
\caption{Quantitative comparison of the automated clustering method K-Means and our manual setting of the threshold $\epsilon$. UI, MB, and UT represent Uni-image, Modality-balance, and Uni-text, respectively.}
\setlength{\tabcolsep}{1.7mm}
\begin{tabular}{@{}c|ccccc|ccccc@{}}
\toprule
       & \multicolumn{5}{c|}{Fakeddit}                 & \multicolumn{5}{c}{MMFakeBench}               \\ \midrule
& UI & MB & UT  & Acc&F1 & UI & MB & UT & Acc&F1 \\ \midrule
Human & 0.15&0.75& 0.10&-&-    &0.13&0.74&0.13&-&-  \\
Ours &0.10[56.67]&0.78[85.53]&0.12[57.14]&\textbf{79.33}&\textbf{64.75}&0.04[69.23]&0.80[85.77]&0.16[75.00]&\textbf{83.33}&\textbf{68.44}   \\
K-Means & 0.11[18.75]&0.86[73.40]&0.03[0.00]&65.33&31.37&0.15[19.57]&0.65[82.56]&0.20[50.85]&66.67&52.94\\
\bottomrule
\end{tabular}
\label{tab:kmeans}
\end{table*}

\subsection{Core Information Extraction}
\label{prompt}
In the view of modality causal effect, we first leverage two large models to extract the core information and then construct the causal graph.
Specifically, we utilize MiniCPM-V~2.6 to identify the core entity $E$ and irrelevant visual content $C$ of images.
Llama3-8B-Instruct is employed to recognize the core word $W$ and the irrelevant word $R$ of texts.
Noted that these large models used for core information extraction do not require further fine-tuning.
The prompts are as follows:

\begin{itemize}
    \item \textbf{MiniCPM-V~2.6}: $<Image>$ Please identify the core entity in this image. Output the corresponding entity region coordinates in the format of [x1, y1, x2, y2], where (x1, y1) denotes the top-left coordinate and (x2, y2) denotes the bottom-right coordinate. Remember to apply coordinate normalization, which means the coordinate range is from 0 to 1.
    \item \textbf{Llama3-8B-Instruct}: Please identify the keyword that can represent the core semantic information of this sentence: $<Text>$. Output the words in the format of [word1, word2, ..., wordn] if the core semantics is word1, word2, ..., and wordn. Please note that the number of words would not be fixed. It depends on your understanding of the sentence.
\end{itemize}

Here we provide some examples (Figure~\ref{fig: exampleOfCore}) to validate the reliability of the extraction results.

To study the effect of different core information extraction models, we adopt additional large models, specifically another LVLM, Ovis1.6-Gemma2-9B~\cite{lu2024ovis}, and another LLM, Yi-1.5-9B~\cite{young2024yi}. Table~\ref{tab:coreOnResults} depicts the ensemble multi-view analysis of different model combinations. Generally, the stronger a large model's reasoning ability, the more accurately it can extract core information. So the overall accuracy of multi-view analysis will be higher. This phenomenon further corroborates the universality and extensibility of the proposed automated analysis. As the capabilities of large models enhance, the accuracy of our proposed automated sample-specific modality bias analysis is anticipated to improve further.

\begin{table*}[h!]
\centering
\caption{The effect of different extraction models on the multi-view analysis. UI, MB, and UT represent Uni-image, Modality-balance, and Uni-text, respectively.}
\setlength{\tabcolsep}{0.8mm}
\begin{tabular}{@{}c|ccccc|ccccc@{}}
\toprule
       & \multicolumn{5}{c|}{Fakeddit}                 & \multicolumn{5}{c}{MMFakeBench}               \\ \midrule
\begin{tabular}[c]{@{}c@{}}Model\\ Combination\end{tabular}  & UI & MB & UT  & Acc&F1 & UI & MB & UT & Acc&F1 \\ \midrule
\begin{tabular}[c]{@{}c@{}}MiniCPM-V 2.6,\\Llama3-8B (Ours)\end{tabular}& 0.10[56.67]&0.78[85.53]&0.12[57.14]&\textbf{79.33}&\textbf{64.75}&0.04[69.23]&0.80[85.77]&0.16[75.00]&\textbf{83.33}&\textbf{68.44} \\
\begin{tabular}[c]{@{}c@{}}MiniCPM-V 2.6,\\ Yi-1.5-9B\end{tabular} & 0.11[50.00]&0.75[87.56]&0.14[48.78]&78.00&62.31& 0.05[0.00]&0.81[79.92]&0.14[71.43]&75.00&52.05\\\midrule
\begin{tabular}[c]{@{}c@{}}Ovis1.6-Gemma2-9B,\\ Llama3-8B\end{tabular} & 0.13[43.59]&0.76[86.03]&0.11[46.88]&76.33&58.55& 0.09[33.33]&0.75[84.89]&0.16[75.00]&78.67&64.78\\
\begin{tabular}[c]{@{}c@{}}Ovis1.6-Gemma2-9B,\\ Yi-1.5-9B\end{tabular} & 0.12[45.95]&0.77[83.98]&0.11[46.88]&75.33 &58.31& 0.04[0.00]&0.82[82.45]&0.14[71.43]&77.33&52.99\\
\bottomrule
\end{tabular}

\label{tab:coreOnResults}
\end{table*}

\subsection{Counterfactual Reasoning}
\label{explanation_counterfactual}
\begin{figure}[t]
  \centering
  \includegraphics[width=\linewidth]{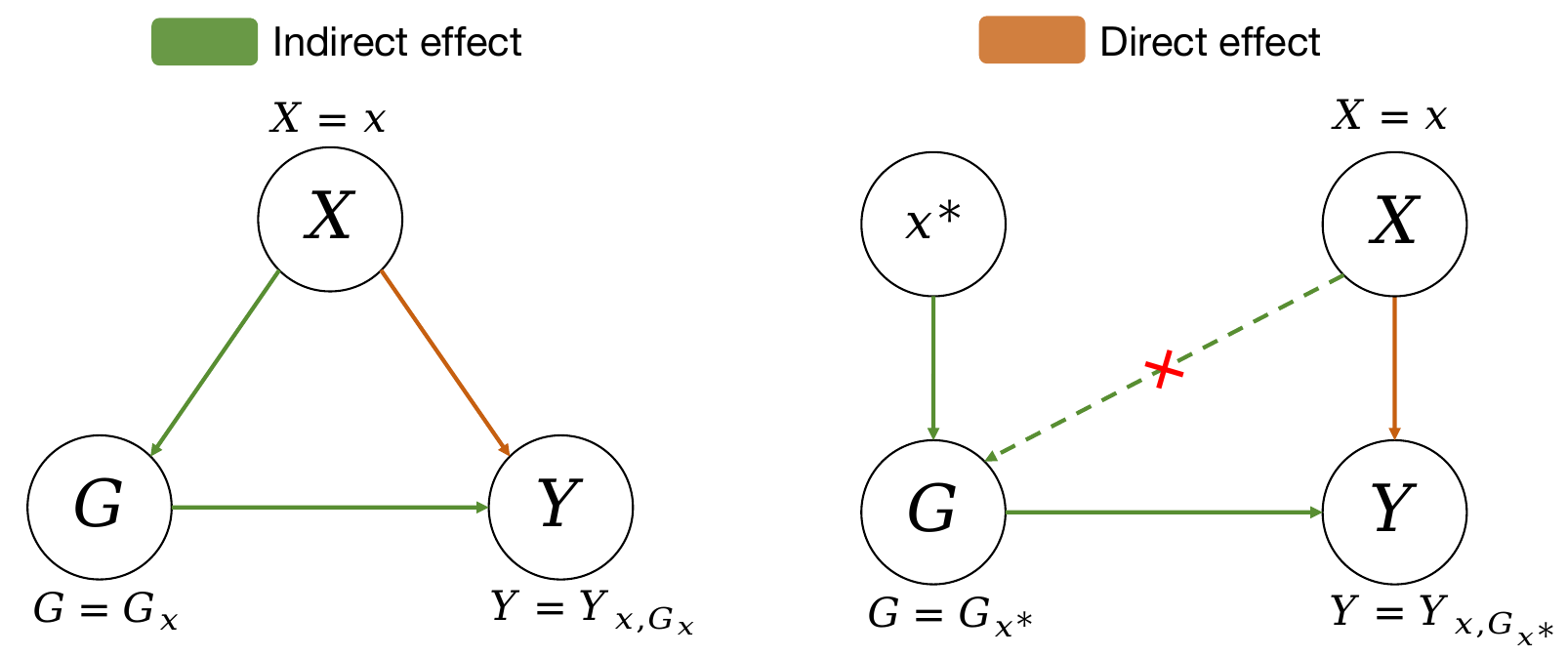}
  \caption{Examples of a causal graph where $X, G, Y$ denote the treatment, mediate, and response variable, respectively.}
  \label{fig: counterfactual}
\end{figure}

Counterfactual analysis~\cite{pearl2009causal} constitutes a fundamental statistical approach for evaluating potential outcomes in hypothetical scenarios that diverge from observed reality. This methodological framework enables researchers to quantify causal relationships between intervention variables and their corresponding outcomes.
For illustration, let us consider a paradigmatic case that contains the treatment variable $X$, the mediate variable $G$, and the response variable $Y$.
Figure~\ref{fig: counterfactual} \textbf{Left} represents the factual scenario, and $Y$ can be mathematically expressed as $Y=Y_{x,G_x}=Y(X=x, G=G(X=x))$, which captures both the direct effect (blown) and indirect effect (green).

When there are two different treatment value of $X$, i.e., $X=x$ and $X=x^*$, the difference of the outcome $Y$, i.e., $Y_{x,G_{x}}-Y_{x^*,G_{x^*}}$ can be seen as the perturbation of $Y$ in response to the change of $X$. Typically, we denote $x^*$ as the reference value and define the
total effect (TE) of $X=x$ on Y as:
\begin{eqnarray}
\text{TE (X on Y)} = Y_{x,G_{x}}-Y_{x^*,G_{x^*}}
\end{eqnarray}

As discussed above, TE contains the direct effect and the indirect effect of $X$ on $Y$.
Now we can use counterfactual reasoning to measure the direct and indirect effect as shown in Figure~\ref{fig: counterfactual} \textbf{Right}. 
Within the counterfactual paradigm, we establish a hypothetical scenario where the treatment variable X simultaneously assumes both $x$ and $x^*$ values. In this constructed setting, the mediator variable $G$ is exclusively determined by $x^*$, while the variable Y receives inputs from both the original treatment value $X=x$ and the mediator $G=G(X=x^*)$. This configuration effectively severs the causal pathway from $X$ to $G$, thereby enabling the analytical separation of the direct and indirect effect.
So the nature direct effect (NDE) of $X$ on $Y$ is:
\begin{eqnarray}
\text{NDE (X on Y)} = Y_{x,G_{x^*}}-Y_{x^*,G_{x^*}}
\end{eqnarray}

Thus, the total indirect effect (TIE) of $X$ on $Y$ can be obtained through:
\begin{eqnarray}
\text{TIE (X on Y)} &=& \text{TE} - \text{NDE} \\ &=& Y_{x,G_{x}} - Y_{x,G_{x^*}}
\end{eqnarray}

Then, to quantize the TE, NDE, and TIE, we need to compute the outputs like $Y_{x,G_{x}}$. Following the previous study \cite{chen2023causal}, we utilize several models and apply a non-linear fusion strategy to obtain such outputs. For example:
\begin{eqnarray}
Y_{x,G_{x}} &=& \mathcal{F}(Y_{x}, Y_{G_x}) \\ &=& tanh(Y_{x}) + tanh(Y_{G_x})
\end{eqnarray}
where $\mathcal{F}(\cdot)$ is the non-linear tanh fusion strategy and the value $Y_x$ is the output probability/logit of corresponding models.

\begin{table}[t]
\centering
\caption{Statistics of the Fakeddit and MMFakeBench.}
\begin{tabular}{@{}cccc@{}}
\toprule
                                 &        & Fakeddit & MMFakeBench \\ \midrule
\multirow{2}{*}{Finetune\_train} & \#Real & 80465    & 1044    \\
                                 & \#Fake & 123281   & 2556    \\ \midrule
\multirow{2}{*}{Finetune\_valid} & \#Real & 8796     & 125     \\
                                 & \#Fake & 13843    & 275     \\ \midrule
\multirow{2}{*}{Analysis\_train} & \#Real & 132820   & 1831    \\
                                 & \#Fake & 204409   & 4169     \\ \midrule
\multirow{2}{*}{Analysis\_valid} & \#Real & 23320    & 300   \\
                                 & \#Fake & 35979    & 700    \\ \midrule
\multirow{2}{*}{Analysis\_test}  & \#Real & 23507    & -      \\
                                 & \#Fake & 35764    & -      \\ \midrule
\multirow{2}{*}{Total}           & \#Real & 268908   & 3300     \\
                                 & \#Fake & 413274   & 7700     \\ \bottomrule
\end{tabular}
\label{tab:Statistics}
\end{table}

\section{Detailed Experiment Design}
\subsection{Statistics of Benchmarks}
\label{Statistics}
Table~\ref{tab:Statistics} depicts the statistics of two multimodal misinformation benchmarks, i.e., Fakeddit and MMFakeBench. Specifically, we report the number of each category (i.e., Real or Fake). Constructed from popular online media, Fakeddit is a highly diverse real-world English benchmark and contains over six hundred thousand multimodal samples.
In contrast, MMFakeBench is a synthetic English dataset generated by Large Vision-language models (LVLM) like DALL-E3. 
For a multimodal misinformation benchmark with a predefined partition of ``Train'', ``Valid'', and ``Test'' sets, we first randomly select 40\% of the samples from the ``Train'' set to fine-tune the models, and then perform sample-specific modality bias analysis on the remaining 60\% of the ``Train'' set, the ``Valid'' set, and the ``Test'' set.
To avoid confusion, we refer to the data used for fine-tuning as ``Finetune\_train'' and ``Finetune\_valid'' 
, while the remaining subsets used for automated analysis are referred to as ``Analysis\_train'', ``Analysis\_valid'', and ``Analysis\_test''.

\subsection{Model Description, Selection Criteria, and Fine-tuning Details}
\label{finetune_detail}

\textbf{Model Description.} We first introduce models utilized in each view.
UnivFD~\cite{ojha2023towards} is a versatile fake image detector that operates within a feature space not explicitly trained to distinguish real from fake images. HAMMER~\cite{shao2023detecting} detects manipulation across different multimedia types. FFNews~\cite{huang2022faking} specializes in detecting textual fake news, particularly human-generated misinformation. MiniCPM-V~2.6~\cite{yao2024minicpm} excels in multimodal understanding. DT($\cdot$)~\cite{papadopoulos2024verite} utilizes CLIP ViT-L/14~\cite{radford2021learning} to extract modality features, with different variants (DT(I), DT(T), DT(I,T)) representing different modality inputs.

\textbf{Model Selection Criteria.} We select these misinformation detection models based on their strong performance and report the detailed quantitative comparison with some other models in Table~\ref{tab:criteria}. 
For \textbf{Image-only} models, we show the performance of Patch classifier~\cite{chai2020makes}, Co-occurence~\cite{nataraj2019detecting} and UnivFD on FaceForensics++~\cite{rossler2019faceforensics++} and LDM~\cite{rombach2022high}.
For \textbf{Image-text} models, we depict the performance of CLIP~\cite{radford2021learning}, ViLT~\cite{kim2021vilt} and HAMMER on DGM4~\cite{shao2023detecting}.
For \textbf{Text-only} models, we compare the performance of DEFEND~\cite{shu2019defend}, DualEmo~\cite{vaibhav2019sentence} and FFNews on PolitiFact~\cite{shu2020fakenewsnet} and LUN~\cite{rashkin2017truth}.
For LVLM, we compare three models of different serials (Ovis1.5-Gemma2-9B~\cite{lu2024ovis}, InternVL2-8B-MPO~\cite{chen2023internvl}, and MiniCPM-V-2.6) and report the average score of eight evaluation datasets (i.e., MMBench~\cite{liu2025mmbench}, MMStar~\cite{chen2024we}, MMMU~\cite{yue2024mmmu}, MathVista~\cite{lu2023mathvista}, AI2D~\cite{kembhavi2016diagram}, HallusionBench~\cite{guan2024hallusionbench}, MMVet~\cite{yu2023mm}, OCRBench~\cite{liu2024ocrbench}) based on VLMEvalKit~\cite{duan2024vlmevalkit}. Note that our framework is adaptable to any misinformation detection method and LVLM.


\textbf{Fine-tuning Details.} Due to the limited performance of existing models in multimodal misinformation detection under zero-shot scenarios, fine-tuning is required for a robust and accurate measurement.
Specifically, we apply supervised fine-tuning (SFT) to UnivFD, HAMMER, FFNews, DT(I), DT(I, T), and DT(T) for 10 epochs.
As for the MiniCPM-V~2.6, we apply LoRA-based parameter-efficient fine-tuning for 1 epoch, considering the balance of resources and accuracy. 
The LVLM(I) indicates that the LVLM exclusively processes image modality inputs, functioning as a unimodal image misinformation detector, while LVLM(I,T) designates the model's operation as a multimodal image-text misinformation detector. Correspondingly, LLM(T) signifies that the LLM operates solely as a text-based misinformation detector.
All hyperparameters are consistent with their original work and experiments are conducted on one A100 80GB GPU.
The accuracy of tuned models on the ``Finetune\_valid'' set is shown in Table~\ref{tab:finetune}.

\begin{table}[t]
\centering
\caption{Quantitative comparison of misinformation detection models and LVLMs.}
\begin{tabular}{@{}cccc@{}}
\toprule
\textbf{Image-only model}& FaceForensics++   & LDM \\ \midrule
Patch classifier           & 75.54     & 79.09  \\
Co-occurence         & 57.10     & 70.70  \\
UnivFD     & \textbf{84.50}     & \textbf{94.19}  \\ \midrule
\textbf{Image-text model} & \multicolumn{2}{c}{DGM4} \\ \midrule
CLIP  & \multicolumn{2}{c}{76.40 }       \\
ViLT        & \multicolumn{2}{c}{78.38}           \\
HAMMER      & \multicolumn{2}{c}{\textbf{86.39}}              \\ \midrule
\textbf{Text-only model}& PolitiFact   & LUN \\ \midrule
DEFEND           & 82.67     & 81.33  \\
DualEmo        & 87.78     & 81.78  \\
FFNews     & \textbf{88.00}     & \textbf{82.53}  \\ \midrule
\textbf{LVLM}& Param (B)   & Avg Score \\ \midrule
Ovis1.5-Gemma2-9B & 11.4     & 64.00  \\
InternVL2-8B-MPO  & 8     & 64.50 \\
MiniCPM-V-2.6     & 8     & \textbf{65.20}  \\ \bottomrule
\end{tabular}
\label{tab:criteria}
\end{table}

\begin{table}[t]
\centering
\caption{The accuracy of tuned models on the ``Finetune\_valid'' set of Fakeddit and MMFakeBench.}
\setlength{\tabcolsep}{0.8mm}
\begin{tabular}{@{}cccc@{}}
\toprule
                           & Model     & Fakeddit  & MMFakeBench \\ \midrule
\multirow{2}{*}{Image-only}&UnivFD     & 79.94     &  74.25      \\
                           &DT(I)      & 88.01     &  80.75      \\ \midrule
\multirow{2}{*}{Image-text}&HAMMER     & 92.41     &  81.00      \\
                           &DT(I, T)   & 93.40     &  83.75      \\ \midrule
\multirow{2}{*}{Text-only} &FFNews     & 89.20     &  86.04      \\
                           &DT(T)      & 88.73     &  75.50      \\ \midrule
                       LVLM (I)&MiniCPM-V 2.6  & 92.71 &  100.00      \\ 
                       LVLM (I,T)&MiniCPM-V 2.6  & 94.61 &  100.00      \\\midrule
                       LLM (T)&Llama3-8B-Instruct  & 95.15 &  100.00      \\\bottomrule
\end{tabular}
\label{tab:finetune}
\end{table}

\subsection{Human Annotation}
\label{human_evaluation}
\citet{liang2024quantifying} shows that human judgment can be used as a reliable estimator of multimodal interaction.
Following their design, we also conduct a human evaluation with three annotators to demonstrate the effectiveness of multi-view analysis.
We recruited the annotators from the local universities of China through public advertisement with a specified pay rate.
They are neither the authors nor members of the authors' research group and are all working towards a graduate degree in computer science, and possess knowledge of multimodal learning.
We pay them 50 CNY an hour.
We show both modalities to the annotators and ask them to annotate the type of modality bias for each sample.
We randomly select 300 samples from each dataset to conduct the experiment.
For Fakeddit, there are 180 samples from ``Analysis\_train'', 60 samples from ``Analysis\_valid'', and 60 samples from ``Analysis\_test''.
For MMFakeBench, there are 180 samples from ``Analysis\_train'' and 120 samples from ``Analysis\_valid''.
For comparison, we report the sample size of the human evaluation in other works, i.e., 50 \cite{liang2024quantifying}; 100 \cite{zhuang2024automatic}; 110 \cite{yan2023gpt}.
We clarify the annotation procedure and judgment criteria before annotation.

\begin{itemize}
    \item Instruction: Given a multimodal news sample, it contains both news caption and news image. You need to rate the following three questions ranging from 0-5.
    \item Question 1. (Uni-Image): The extent to which \textbf{Image} modality enables you to predict without the other modality.
    \item Question 2. (Uni-Text): The extent to which \textbf{Text} modality enables you to predict without the other modality.
    \item Question 3. (Modality-balance): The extent to which \textbf{both} modalities enable you to predict that you would not otherwise make using either modality individually.
\end{itemize}

For a specific sample, we first average the three scores of each annotator, respectively, and then select the type with the highest score as the bias type of this sample.

We conducted Krippendorff's alpha test~\cite{krippendorff2011computing} to assess the reliability and agreement of human annotations. As presented in Table~\ref{tab:agreementTest}, all alpha values exceed 0.8, which demonstrates a high level of agreement among the three annotators and further substantiates the validity of our human annotations.

\begin{table}[t]
    \centering
    \caption{The Krippendorff's alpha test of human annotations.}
    \setlength{\tabcolsep}{1.8mm}
    \begin{tabular}{@{}cccc@{}}
    \toprule
        ~ & Uni-image & Modality-balance & Uni-text \\ \midrule
        Fakeddit & 0.8105 & 0.8547 & 0.8752 \\ 
        MMFakeBench & 0.8017 & 0.8557 & 0.8375 \\ \bottomrule
    \end{tabular}
    \label{tab:agreementTest}
\end{table}

\section{Extend to Large Models}
\label{extend_to_large_models}
Recent advances have demonstrated the superior performance of large models across diverse tasks. Motivated by these developments, we conduct experimental investigations into migrating our automated analysis framework to large model-based architectures. Notably, our automated analysis exhibits remarkable transferability, as it only requires different prompts to adapt large models to different model categories (i.e., Image-only, Image-text, and Text-only models). Here we utilize MiniCPM-V 2.6 as the Image-only (LVLM(I)) and Image-text (LVLM(I,T)) model, and LLama3-8B-Instruct as the Text-only model. The prompts are  as follows:
\begin{itemize}
    \item \textbf{LVLM(I)}: $<Image>$ Question: Given a news image, is there any misinformation in the news? Options: 0. No. 1. Yes. 
    \item \textbf{LVLM(I, T)}: $<Image>$ Given a multimodal news, it contains both the news caption and the news image. The following is a multiple-choice question about multimodal fake news detection. There is no relationship between the answer and options' position. The caption of the news is: $<Text>$. Question: Considering both the image and the caption, is there any misinformation in the news? Options: 0. No. 1. Yes.
    \item \textbf{LLM(T)}: The caption of the news is: $<Text>$. Question: Is there any misinformation in the news? Options: 0. No. 1. Yes.
\end{itemize}

Table~\ref{tab:large_models_result} depicts the automated analysis with pure large models, i.e., \textbf{Image-only:} LVLM(I); \textbf{Image-text:} LVLM(I,T); \textbf{Text-only:} LLM(T). 
Our experimental observations reveal several anomalous phenomena: (1) In the ablation study, both Benefit-Flow and Benefit-Causal unexpectedly outperformed the Multi-view analysis; (2) On MMFakeBench, Modality Benefit, Benefit-Flow, and Benefit-Causal all exhibited severe distributional bias by classifying every sample as Modality-balance. From Table~\ref{tab:large_models_result}, there is a counterintuitive finding that automated analysis based on pure large models underperformed compared to its smaller counterpart.

Because the anomalous results all contain the Modality Benefit view, we conducted a systematic analysis on the computational framework and the fine-tuning results of large models presented in Table~\ref{tab:finetune}, we identify a \textbf{saturation effect}: the large models achieved 100\% validation accuracy across all three mode (LVLM(I), LVLM(I,T), and LLM(T)) on MMFakeBench. This over-perfect performance renders the Modality Benefit view theoretically degenerate, as the underlying assumption of measurable performance differences between modalities no longer holds when all modalities achieve maximal accuracy. In other words, for any situations, we have $V\left ( x^{image},0^{text} \right )=1$, $V\left ( x^{text},x^{image} \right )=2$, and $V\left ( x^{text},0^{image} \right )=1$. Therefore, $\phi _{m_{image}} = \phi _{m_{text}}$ always holds true. That is why automated analysis based on pure large models performs worse than its smaller counterparts.

\begin{table*}[t]
\centering
\caption{The performance of automated analysis with pure large models. UI, MB, and UT represent Uni-image, Modality-balance, and Uni-text, respectively.}
\setlength{\tabcolsep}{0.9mm}
\begin{tabular}{@{}c|ccccc|ccccc@{}}
\toprule
       & \multicolumn{5}{c|}{Fakeddit}                 & \multicolumn{5}{c}{MMFakeBench}               \\ \midrule
       & UI & MB & UT  & Acc&F1 & UI & MB & UT & Acc&F1 \\ \midrule
Human          &0.15&0.75& 0.10&-&-    &0.13&0.74&0.13&-&-       \\
Modality benefit &0.03[0.00]&0.84[76.10]&0.13[37.50]&68.67&41.04&0.00[0.00]&1.00[73.67]&0.00[0.00]&73.67&28.28   \\
Modality flow &0.13[42.50]&0.52[96.15]&0.35[28.85]&65.67&54.51&0.04[0.00]&0.67[70.00]&0.39[12.50]&50.33&27.85   \\
Modality causal effect &0.08[0.00]&0.88[71.59]&0.04[0.00]&63.00&25.77&0.02[0.00]&0.78[76.82]&0.20[40.32]&68.00&42.47   \\
Multi-view analysis &0.03[0.00]&0.87[76.92]&0.10[48.39]&71.67&43.88&0.00[0.00]&0.98[73.04]&0.02[0.00]&71.33&27.76   \\
\midrule
Benefit-Flow   &0.03[0.00]&0.90[77.86]&0.07[75.00]&75.33&48.36&0.00[0.00]&1.00[73.67]&0.00[0.00]&73.67&28.28    \\
Benefit-Causal &0.00[0.00]&0.98[74.58]&0.02[0.00]&73.33&28.21&0.00[0.00]&1.00[73.67]&0.00[0.00]&73.67&28.28    \\
Flow-Causal    &0.06[0.00]&0.92[72.83]&0.02[0.00]&67.00&26.75&0.00[0.00]&0.98[73.04]&0.02[0.00]&71.33&27.76    \\
\bottomrule
\end{tabular}
\label{tab:large_models_result}
\end{table*}

\begin{table*}[]
\centering
\caption{Post-validation of our automated analysis. We report the accuracy (without []) and the difference (within []) compared to the condition without modification, i.e., (Image, Text): (0.0, 0.0).}
\setlength{\tabcolsep}{5mm}
\begin{tabular}{@{}c|cc|cc@{}}
\toprule
Mask ratio & \multicolumn{2}{c|}{Fakeddit} & \multicolumn{2}{c}{MMFakeBench}\\ 
(Image, Text) & Uni-image & Uni-text & Uni-image & Uni-text \\ \midrule
(0.6,0.0) & 77.00[\textbf{-5.22}] & 85.46[-5.06] & 69.92[\textbf{-3.64}] & 60.00[-6.12]\\
(0.4,0.0) & 79.39[-2.83] & 87.15[-3.37] & 72.41[-1.15] & 63.27[-2.85]\\
(0.2,0.0) & 81.22[-1.00] & 88.59[-1.93] & 72.99[-0.57] & 64.90[-1.22]\\
(0.0,0.0) & \textbf{82.22} & \textbf{90.52} & \textbf{73.56} & \textbf{66.12}\\
(0.0,0.2) & 81.88[-0.34] & 88.31[-2.21] & 73.18[-0.38] & 55.84[-10.28]\\
(0.0,0.4) & 81.38[-0.84] & 85.25[-5.27] & 72.80[-0.76] & 44.65[-21.47]\\
(0.0,0.6) & 80.40[-1.82] & 81.25[\textbf{-9.27}] & 72.41[-1.15] & 37.55[\textbf{-28.57}]\\ \bottomrule
\end{tabular}%
\label{tab:post_validation}
\end{table*}

\begin{table}[]
\centering
\caption{The accuracy [\%] of modality benefit, modality causal effect, and multi-view analysis on Fakeddit with ensembling multiple misinformation detectors .}
\setlength{\tabcolsep}{1.8mm}
\begin{tabular}{c|ccc}
\toprule
        &\begin{tabular}[c]{@{}c@{}}Modality\\ benefit\end{tabular}&\begin{tabular}[c]{@{}c@{}}Modality\\ causal effect\end{tabular} &\begin{tabular}[c]{@{}c@{}}Multi-view\\ analysis\end{tabular}  \\ \midrule
Ours & 71.67  & 64.67  & 79.33  \\
Multi-Detectors& 71.67  & 62.00  & 73.33  \\
\bottomrule
\end{tabular}
\label{tab:multi_detectors}
\end{table}

\section{Error Analysis}
\label{error_analysis}
\begin{figure}[t]
  \centering
  \includegraphics[width=\linewidth]{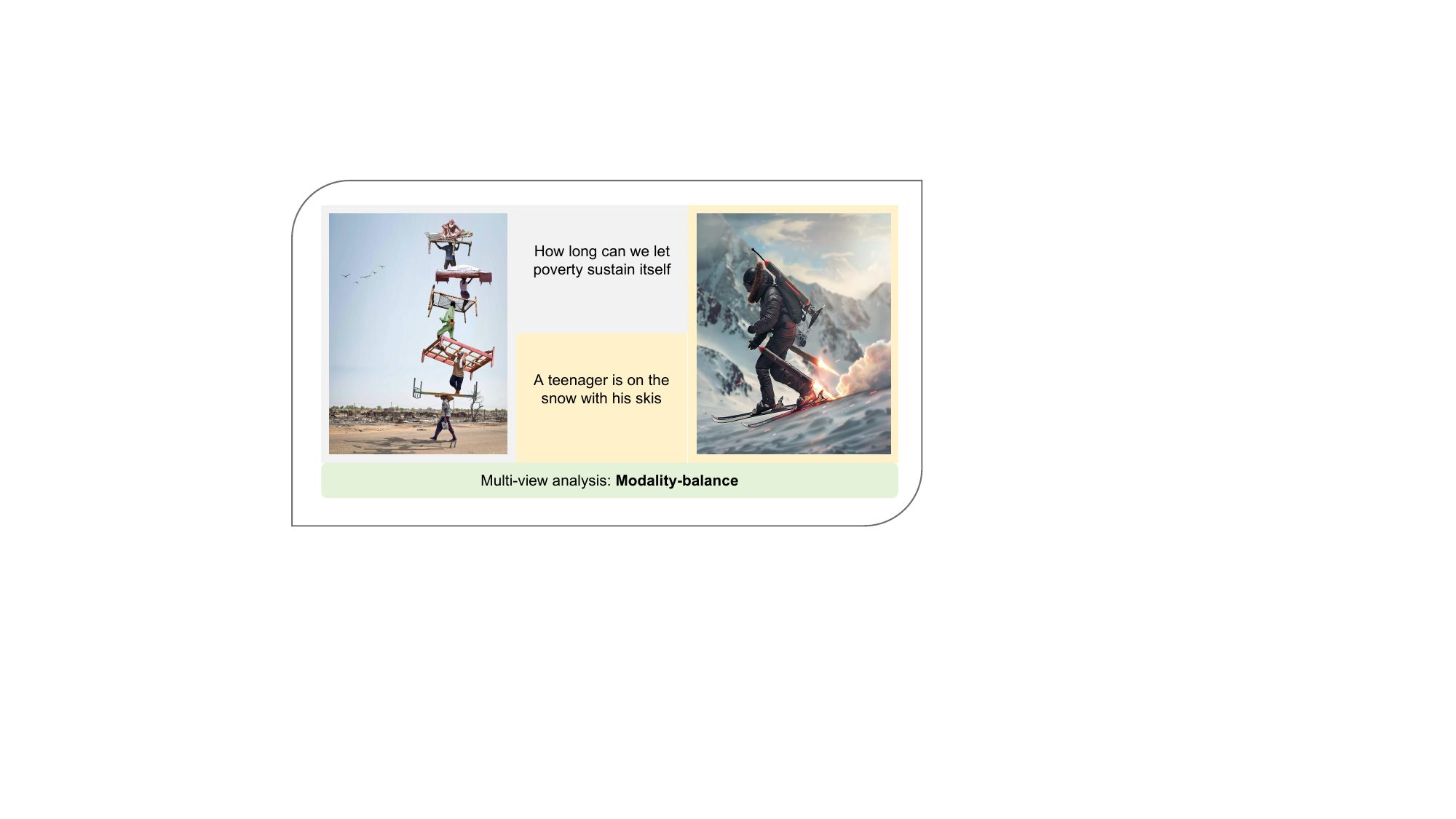}
  \caption{Error cases of multi-view analysis. The modality bias of these two samples should be ``Uni-image''.}
  \label{fig: error}
\end{figure}

As shown in Figure~\ref{fig: error}, we conduct an error analysis on the ``Uni-image'' category, which exhibited the lowest performance in our multi-view analysis. We found that the multi-view analysis struggles to correctly identify well-edited images (Figure~\ref{fig: error}, left) or images synthesized by large vision-language models (Figure~\ref{fig: error}, right). Although these images may appear seamless at the pixel level, they contain misinformation at the semantic level. However, the multi-view analysis incorrectly classifies these samples as ``Modality-balance''. We attribute this issue to the limitations of current MMD models, which are not yet equipped to handle such complex cases. As more advanced techniques are developed, these types of errors may decrease, improving the accuracy of automated bias evaluation systems.

\section{Post-validation of automated analysis}
\label{post_validation}
Here, we provide an alignment test between our automated analysis and the behaviors of misinformation detector models, i.e., the modality-specific perturbation effects on model performance, with particular focus on samples classified as modality-biased (Uni-image and Uni-text) by our automated analysis.
Specifically, we conduct an experiment comparing the \textbf{multimodal misinformation detection results} of the HAMMER under different modifications on the whole dataset of Fakeddit and MMFakeBench, where we mask the input image or text to a certain ratio. We report the accuracy (without []) and the difference (within []) compared to the condition without modification (mask ratio=0.0) on the entire Fakeddit and MMFakeBench in Table~\ref{tab:post_validation}. The results indicate that biased samples exert a greater influence on the model when subjected to perturbations from the corresponding modality, which further demonstrates the effectiveness of our automated analysis.

\section{Ensemble Multiple Detectors}
\label{multi_detectors}
Since the computational process of our automated analysis fundamentally relies on models' outputs, its performance is inherently susceptible to variations across different misinformation detectors. To explore the potential solution, we investigate whether ensembling multiple misinformation detectors can enhance the robustness and stability of our automated analysis. Specifically, we integrate three detectors per model category on Fakeddit, i.e., \textbf{Image-only:} UnivFD, DT(I), and LVLM(I); \textbf{Image-text:} HAMMER, DT(I,T), LVLM(I,T); \textbf{Text-only:} FFNews, DT(T), and LLM(T). The detailed description of these models can be found in Appendix~\ref{finetune_detail}. We utilize MiniCPM-V 2.6 as the Image-only (LVLM(I)) and Image-text (LVLM(I,T)) model, and LLama3-8B-Instruct as the Text-only model. Since the performance of the large models-driven \textbf{multi-view} analysis on Fakeddit is reasonable, we incorporate large models-based misinformation detectors into this investigation.
The prompts are the same as those in Appendix~\ref{extend_to_large_models}.
The quantitative result of ensembling multiple detectors (Multi-Detectors) is shown in Table~\ref{tab:multi_detectors}, which demonstrates enhancements in the robustness of our automated analysis when compared to the performance fluctuation using a single detector for per category.

\end{document}